\documentclass{bmvc2k}

\title{Self-Validation: Early Stopping for Single-Instance Deep Generative Priors}

\addauthor{Taihui Li}{lixx5027@umn.edu}{1}
\addauthor{Zhong Zhuang}{zhuan143@umn.edu}{2}
\addauthor{Hengyue Liang}{liang656@umn.edu}{2}
\addauthor{Le Peng}{peng0347@umn.edu}{1}
\addauthor{Hengkang Wang}{wang9881@umn.edu}{1}
\addauthor{Ju Sun}{jusun@umn.edu}{1}

\addinstitution{
 Department of Computer Science and Engineering\\
 University of Minnesota\\
 Minnesota, USA
}
\addinstitution{
 Department of Electrical and Computer Engineering\\
 University of Minnesota\\
 Minnesota, USA
}

\runninghead{Li et al.}{Self-Validation: Early Stopping for SIDGPs}

\usepackage{algorithm}
\usepackage{algorithmic}

\usepackage{graphicx,amsmath,amsfonts,amsthm,amscd,amssymb,bm,url,color,latexsym}
\usepackage{physics}
\allowdisplaybreaks
\usepackage[capitalize,nameinlink]{cleveref}

\renewcommand{\mathbf}{\boldsymbol}

\newcommand{\mb}{\mathbf}
\newcommand{\mc}{\mathcal}

\newcommand{\bb}{\mathbb}

\newcommand{\set}[1]{\left\{ #1 \right\}}

\newcommand{\eps}{\varepsilon}
\newcommand{\R}{\bb R}

\newcommand{\N}{\bb N}

\newcommand{\paren}{\pqty}

\newcommand{\wh}{\widehat}

\newcommand{\ol}{\overline}

\newcommand{\T}{\intercal}

\newcommand{\expect}[1]{\bb E\left[ #1 \right]}

\usepackage{subfigure}
\usepackage{float}
\usepackage{wrapfig}
\usepackage{booktabs}
\usepackage{multirow}
\usepackage{makecell}
\usepackage{enumitem}
\newcommand{\tabincell}[2]{\begin{tabular}{@{}#1@{}}#2\end{tabular}}

\usepackage{threeparttable}

\begin{document}

\maketitle

\begin{abstract}

Recent works have shown the surprising effectiveness of deep generative models in solving numerous image reconstruction (IR) tasks, \emph{even without training data}. We call these models, such as deep image prior and deep decoder, collectively as \emph{single-instance deep generative priors} (SIDGPs). The successes, however, often hinge on appropriate early stopping (ES), which by far has largely been handled in an ad-hoc manner. In this paper, we propose the first principled method for ES when applying SIDGPs to IR, taking advantage of the typical bell trend of the reconstruction quality. In particular, our method is based on collaborative training and \emph{self-validation}: the primal reconstruction process is monitored by a deep autoencoder, which is trained online with the historic reconstructed images and used to validate the reconstruction quality constantly. Experimentally, on several IR problems and different SIDGPs, our self-validation method is able to reliably detect near-peak performance and signal good ES points. Our code is available at~\url{https://sun-umn.github.io/Self-Validation/}.

\end{abstract}

\section{Introduction}
\label{sec:intro}
Validation-based ES is one of the most reliable strategies for controlling generalization errors in supervised learning, especially with potentially overspecified models such as in gradient boosting and modern DNNs~\cite{zhang2005boosting,prechelt1998early,Goodfellow-et-al-2016}. Beyond supervised learning, ES often remains critical to learning success, but there are no principled ways---universal as validation for supervised learning---to decide when to stop. In this paper, we make the first step toward filling in the gap, and focus on solving IR, a central family of inverse problems, using \emph{training-free deep generative models}.

\begin{wraptable}{r}{0.38\linewidth}
  \centering
  \vspace{-0.5em}
  \resizebox{\linewidth}{!}{
   \begin{tabular}{l  c}
      \multicolumn{2}{c}{\textbf{List of Acronyms} (alphabetic order)} \\
      \hline
      AE  & Autoencoder \\
      CNN  & convolutional neural network \\
      DD  & deep decoder \\
       DGP & deep generative prior \\
        DIP  & deep image prior \\
      DL & deep learning \\
        DNN  & deep neural network \\
      ES & early stopping \\
      IR  & image reconstruction \\
      MIDGP  & multi-instance DGP \\
      OPT & optimization \\
      PG  & PSNR gap \\
       SG  & SSIM gap \\
      SIDGP  & single-instance DGP \\
      SIREN  & sinusoidal representation network \\
      \hline
    \end{tabular}
    }
\end{wraptable}IR entails estimating an image of interest, denoted as $x$, from a measurement $y = f\paren{x}$, where $f$ models the measurement process. This model covers classical image processing tasks such as image denoising, superresolution, inpainting, deblurring, and modern computational imaging problems such as MRI/CT reconstruction~\cite{epstein2007introduction} and phase retrieval~\cite{shechtman2015phase,TayalEtAl2020Inverse}. In this paper, \emph{we assume that $f$ is known}. Classical approaches normally formulate IR as a regularized data fitting problem (Ref. problem \eqref{eq:inv_obj_1}) and solve it via iterative optimization algorithms~\cite{parikh2014proximal}. Recently, DL-based methods have been developed to either directly approximate the inverse mapping $f^{-1}$, or enhance classical optimization algorithms for solving problem \eqref{eq:inv_obj_1} by integrating pretrained or trainable DNN modules (e.g., plug-and-play or network unrolling; see the recent survey~\cite{ongie2020deep}). But, they invariably require extensive and representative training data. In this paper, we focus on an emerging lightweight approach \emph{that requires no extra training data}: the target $x$ is directly modeled via DNNs. 
\begin{align}  \label{eq:inv_obj_1}
    \min_x\; \ell\paren{y, f\paren{x}} + \lambda R\paren{x} \;\; \ell\text{: fitting loss},\;  R\text{: regularization},\;  \lambda\text{: regularization parameter} 
\end{align}

\begin{figure}[!htpb]
\centering
\vspace{-2em}
\includegraphics[width=0.8\textwidth]{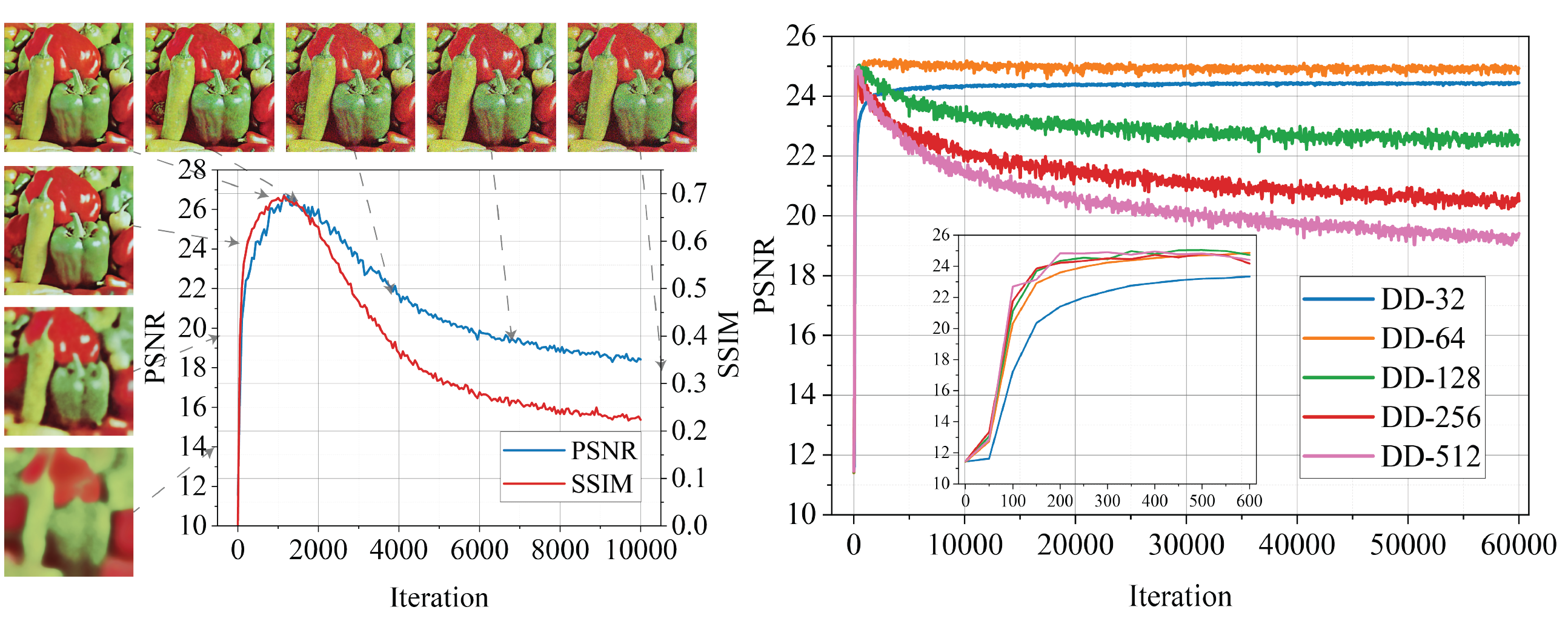}
\vspace{-1em}
\caption{Illustration of the overfitting issue of DIP and DD on image denoising with Gaussian noise (noise level: $\sigma = 0.18$; pixel values normalized to $[0, 1]$). The reconstruction quality (as measured by both PSNR and SSIM) typically follows a skewed bell curve: before the peak only the clean image content is recovered, but after the peak noise starts to kick in. Note that DD is not free from overfitting when the network is increasingly over-parametrized (indicated by the number following ``DD-'' in the legend). }
\label{fig:dip_dd_overfit_curve}
\end{figure}

\textbf{Single-instance deep generative priors} \quad
Many deep generative models can be used for modeling image collections~\cite[Part III]{Goodfellow-et-al-2016}. For modeling single images, two families of models stand out: 1) \emph{structural models}\footnote{Some authors call this \emph{untrained DNN models/priors}; see~\cite{qayyum2021untrained}. But this may be confused with SIDGPs with prefixed but frozen $G$ (e.g., random) as in~\cite{bora2017compressed,hand2018phase}.  }: the image is modeled as $x \approx G_{\theta}\paren{z}$. Here, $z$ is a fixed or trainable seed/code, and $G$ is a trainable DNN---often taken as CNN to bias toward structures in natural images---parameterized by $\theta$. DIP~\cite{ulyanov2018deep} and DD~\cite{heckel2018deep} are two exemplars in this category, and they only differ in the choice of architecture for $G$. A variant~\cite{pan2020exploiting} allows initializing $G$ with a pretrained model. These models should be contrasted with \emph{multi-instance DGPs} (MIDGPs, of which GAN inversion~\cite{zhu2016generative,bau2019seeing} is a special case), where $G$ is pretrained and frozen but $z$ is trainable. Although MIDGPs have also been used to tackle IR problems~\cite{bora2017compressed,hand2018phase}, pretraining $G$ requires massive training sets that can cover the target domains. Thus, MIDGPs are not suitable for single-image settings\footnote{See~\cite{leong2019low} that tries to bridge the two regimes though. }. 2) \emph{functional models}: the (discrete) image $x$ is modeled as the collection of uniform-grid samples, i.e., discretization, from an underlying \emph{continuous} function $\ol{x}$ supported on a bounded region in $\R^2$ (sometimes higher dimensions)\footnote{In a sense, this is inverting the discretization process in typical image formation. Also, the idea of modeling continuous functions directly is popular in DL for solving PDEs~\cite{weinan2020machine}. }. Now $\ol{x}$ is parameterized as a DNN $\ol{x}_\theta$, i.e., $x = \mc S\paren{\ol{x}_\theta}$ where $\mc S$ denotes the sampling process. Note that if $\ol{x}$ can be learned, in principle one can obtain arbitrary resolutions for the discrete version $x$---attractive for computer vision/graphics applications. These models are called implicit neural representation or coordinate-based multilayer perceptron (MLP) in the literature, of which sinusoidal representation networks (SIRENs,~\cite{sitzmann2020implicit}) and MLP with Fourier features~\cite{tancik2020fourfeat} are two representative models.

We categorize all these single-image models into the family of SIDGPs. Substituting any of the SIDGPs into problem~\eqref{eq:inv_obj_1} and removing the explicit regularization term $R$, we can now solve IR problems by DL:
\begin{align}  \label{eq:inv_obj_2}
    \textbf{structural OPT}: \; \min_{\theta}\; \ell\paren{y, f\paren{G_\theta\paren{z}}}   \quad \text{or} \quad   \textbf{functional OPT}: \; \min_{\theta}\; \ell\paren{y, f\paren{\mc S\paren{\ol{x}_\theta}}}.
\end{align}
This simple approach is surprisingly effective in solving numerous IR problems, ranging from classical image restoration~\cite{ulyanov2018deep,heckel2018deep}, to advanced computational imaging problems~\cite{bora2017compressed,hand2018phase,gandelsman2019double,ren2020neural, williams2019deep,sitzmann2020implicit,tancik2020fourfeat,darestani2020accelerated}, and even beyond~\cite{ravula2019one,michelashvili2019audio}; see the recent survey~\cite{qayyum2021untrained}.

\textbf{The overfitting issue} \quad 
There is a caveat to all the claimed successes: overfitting. In practice, we have $y \approx f\paren{x}$ instead of $y = f\paren{x}$ due to various kinds of measurement noise and model misspecification, and the DNN used in \eqref{eq:inv_obj_2} is also often (substantially) overparametrized. So when the objective in \cref{eq:inv_obj_2} is globally optimized, the final reconstruction $G_\theta\paren{z}$ or $\mc S\paren{\ol{x}_\theta}$ may account for the noise and model errors, alongside the desired image content. This overfitting phenomenon has indeed been observed empirically on all SIDGPs when noise is present. \cref{fig:dip_dd_overfit_curve} shows the typical reconstruction quality of DIP and DD for image denoising (Gaussian noise) over iterations. Note the iconic skewed bell curves here: the reconstruction quality initially monotonically climbs to a peak level---noise effect is almost invisible during this period, and then monotonically degrades where the noise effect gradually kicks in until everything is fit by the overparametrized models. Also, when the level of overparametrization grows, overfitting becomes more serious (\cref{fig:dip_dd_overfit_curve} (b) for DD). The overfitting phenomenon has also been partly justified theoretically: despite the typical nonconvexity of the objectives in \cref{eq:inv_obj_2}, global optimization is feasible and happens with high probability~\cite{jagatap2019algorithmic,heckel2019denoising}. But it is a curse here. 

\textbf{Prior work on mitigating overfitting} \quad 
Several lines of methods have been developed to remove the overfitting. DD~\cite{heckel2018deep} proposes to control the network size for structural models, and shows that overfitting is largely suppressed when the network is suitably parameterized. However, choosing the right level of parameterization for an unknown image with unknown complexity is no easy task, and underparameterization can apparently hurt the performance, as acknowledged by~\cite{heckel2018deep}. So in practice people still tend to use overparametrization and counter overfitting with ES~\cite{pmlr-v119-heckel20a}; see~\cref{fig:dip_dd_overfit_curve} (b). The second line adds regularization. \cite{sun2020solving,sun2021plug} plug in total-variation or other denoising regularizers, and \cite{cheng2019bayesian} appends weight-decay and runs SGD with Langevin dynamics (i.e., with additional noise)~\cite{welling2011bayesian}. But their evaluations are limited to denoising with low Gaussian noise; our experiments in \cref{subsec:image_denoising} show that overfitting still exists when the noise level is higher. The third line explicitly models the noise in the objective, so that hopefully both the image $x$ and the noise are exactly recovered when the objective is globally optimized. \cite{NEURIPS2020_cd42c963} implements such an idea for additive sparse corruption. The overfitting is gone, but it is unclear how to generalize the algorithm, which is delicately designed for sparse corruption only, to other types of noise.

\textbf{Our contribution} \quad 
In this paper, we take a different route, and capitalize on overfitting instead of suppressing or even killing it. Specifically, we leave the problem formulation~\eqref{eq:inv_obj_2} as is, and propose a reliable method for detecting near-peak performance, which is \emph{first of its kind} and also 1) \textbf{effective}: empirically, our method detects near-peak performance measured in both PSNR and SSIM; 2) \textbf{fast}: our method performs ES once a near-peak performance is detected---which is often early in the optimization process (see~\cref{fig:dip_dd_overfit_curve}), whereas the above mitigating strategies push the peak performance to the final iterations; 3) \textbf{versatile}: the only assumption that our method relies on is the performance curve (either PSNR or SSIM) is (skewed) bell-shaped. This seems to hold for the various structural and functional models, tasks, noise types that we experiment with; see~\cref{sec:exps}. 

After the initial submission, we became aware of two new works~\cite{ShiEtAl2021Measuring,JoEtAl2021Rethinking} that also perform ES. \cite{ShiEtAl2021Measuring} proposes an ES criterion based on a blurriness-to-sharpness ratio for natural images, but it only works for the modified DIP and DD they propose, not the original and other SIDGPs. \cite{JoEtAl2021Rethinking} focuses on Gaussian noise, and designs an elegant Gaussian-specific regularized objective to monitor progress and decide ES. Compared to our approach, both methods are limited in versatility. We defer a detailed comparison with them to future work.


\section{Early stopping via self-validation}
\label{sec:method}

\begin{figure}[!htpb]
\vspace{-2em}
\centering
{\includegraphics[width=0.8\textwidth]{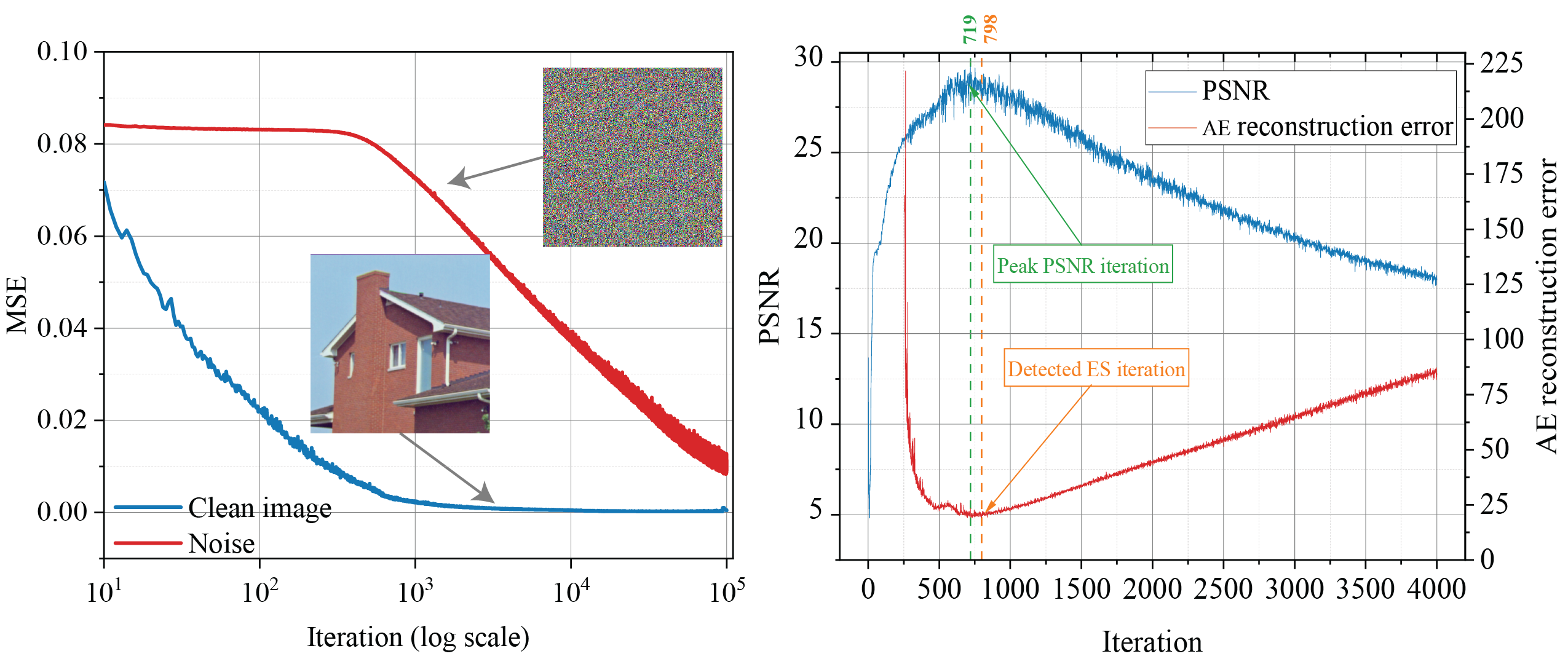}} 
\vspace{-1em}
\caption{(left) The MSE curves of learning a natural image vs learning random noise by DIP. DIP fits the natural image much faster than noise. This induces the typical bell-shaped reconstruction quality curves (as shown in \cref{fig:dip_dd_overfit_curve} and (right) here) when fitting noisy images using DIP; (right) The PSNR curve vs our online AE reconstruction error curve when fitting a noisy image with DIP. The peak of the PSNR curve is well aligned with the valley of the AE error curve. So by detecting the valley of the latter curve, we are able to detect near-peak points of the former---corresponding to reconstructed images of near-peak quality. }
\label{fig:SV_illus}
\end{figure}
In this section, we describe our detection method taking structural SIDGPs as an example; this can be easily adapted for functional SIDGPs. 

\textbf{Bell curves: the why and the bad} \quad
When $y = f\paren{x}$, solving the structural OPT in~\eqref{eq:inv_obj_2} with an overparameterized CNN $G_\theta$ using gradient descent has disparate behaviors on natural images vs pure noise: the term $G_{\theta}\paren{z}$ fits natural images much more rapidly than fitting pure noise. This is illustrated in \cref{fig:SV_illus} (left) and has been highlighted in numerous places~\cite{heckel2018deep,heckel2019denoising}; it can be explained by the strong bias of convolutional operations toward natural image structures, which are low-frequency dominant, during the early learning stage~\cite{heckel2018deep,heckel2019denoising}. When $y \approx f\paren{x}$, e.g., the measurement is noisy, intuitively $G_{\theta}\paren{z}$ would quickly fit the image content with little noise initially, and then gradually fit more noise until reaching the full capacity. This induces the skewed bell curves as shown in \cref{fig:dip_dd_overfit_curve} and \cref{fig:SV_illus} (right). Of course, this argument is handwavy and the fitting effect might not be decomposable into that of image content and noise separately for complicated $f$'s and noise types (it has been rigorously established for simple cases, e.g., ~\cite{heckel2019denoising}). Nonetheless, the bell-shaped performance curves are also observed for nonlinear $f$'s with complex noise types,  e.g.,~\cite{bostan2020deep,lawrence2020phase}, as well as functional models as we show in~\cref{subsec:img_reg}. The downside is overfitting, if no suitable ES is performed. Recent methods that mitigate the overfitting issue defer the performance peak until the final convergence and work only in limited scenarios. 

\textbf{Bell curves: the good} \quad 
Here we do not try to alter the bell curves but turn them into our favor, as explained below. In practice, we do not directly observe these curves, as they depend on the groundtruth $x$. Let $\set{\theta_k}$ denote the iterate sequence and $\set{x_k}$ the corresponding reconstruction sequence, i.e., $x_k = G_{\theta_k}\paren{z}$. We only observe $\set{x_k}$ and $\set{\theta_k}$ directly, and we know that there exists a $k^* \in \N$ so that the subsequence $\set{x_k}_{k \le k^*}$ has increasingly better quality and contains very little noise. 

Our first technical idea is constructing a quality oracle by training an AE
\begin{align} \label{eq:ae_form}
    \min_{w, v} \; \sum_{i=1}^n \ell_{\mathrm{AE}} \paren{s_i, d_{v} \circ e_{w} \paren{s_i}} \quad e_w\text{: encoder DNN}\quad  d_v\text{: decoder DNN}, 
\end{align}
where $\set{s_i}_{i \le n}$ is a given training set. AEs are now a seminal tool for manifold learning and nonlinear dimension reduction~\cite[Chap. 14]{Goodfellow-et-al-2016}. When $\set{s_i}_{i \le n}$ is sampled from a low-dimensional manifold, after training we expect that $s \approx d_v \circ e_w\paren{s}$ for any $s$ from the same manifold and $s \not\approx
 d_v \circ e_w\paren{s}$ otherwise. This implies that we can use the AE reconstruction loss $\ell_{\mathrm{AE}} \paren{s, d_{v} \circ e_{w} \paren{s}}$ as a proxy to tell if a novel data point $s$ comes from the training manifold. Thus, if the training set $\set{s_i}_{i \le n}$ consists of images very close to $x$, we can use $\ell_{\mathrm{AE}} \paren{s, d_{v} \circ e_{w} \paren{s}}$ to score the reconstruction quality of any $s$: higher the loss, lower the quality, and vice versa. If we apply this ideal AE to $\set{x_k}$, the AE loss curve will have an inverted bell shape, where the valley corresponds to the quality peak. So we can detect the peak of the quality curve by detecting the valley of the AE loss curve. 
 
But it is unclear how to ensure that $\set{s_i}_{i \le n}$ be uniformly close to $x$. Our second technical idea is exploiting the monotonicity of reconstruction quality in $\set{x_k}_{k \le k^*}$ and $\set{x_k}_{k \ge k^*}$ to train a sequence of AEs. We take a consecutive length-$n$ subsequence of $\set{x_k}_{k \le k^*}$ to train an AE each time, and test the next reconstructed image against the resulting AE. Although the quality of any initial training set can be low, resulting in poor AEs, it improves over time and culminates when getting near the $x_{k^*}$. Similarly, afterward, the quality gradually degrades. Away from the peak $x_{k^*}$, the quality of $x_k$'s changes rapidly over iterations, and so we expect high AE loss, regardless of the quality of the AE training set. In contrast, around $x_{k^*}$ all $x_k$'s are of high quality, leading to near-ideal AEs and low AE losses. Thus, even after the practical modification, we expect an inverted bell shape in the AE loss curve, and alignment of the quality peak with the loss valley again, as confirmed in \cref{fig:SV_illus} (right). 

\begin{algorithm}[!htbp]
\footnotesize
\caption{Structural OPT with ES for IR}
\label{alg:framework} 
\begin{algorithmic}[1]
\REQUIRE $y, f, \ell, \ell_{\mathrm{AE}}$, $G$, $\theta^{(0)}$ (of \cref{eq:inv_obj_2}), $w^{(0)}$, $v^{(0)}$ (of \cref{eq:ae_form}), $k = 0$, window size $n$, patience number $p$
\WHILE{not stopped}
\STATE update $\theta$ via one iterative step on \cref{eq:inv_obj_2} to obtain $\theta^{(k+1)}$ and $x^{(k+1)}$
\STATE update $w, v$ via one iterative step on \cref{eq:ae_form} using $\set{x_i}_{i = k-n+1}^{k}$ to obtain $w^{(k+1)},v^{(k+1)}$
\STATE calculate the reconstruction loss $\ell_{\mathrm{AE}} (x^{(k+1)}, d_{v^{(k+1)}} \circ e_{w^{(k+1)}} (x^{(k+1)}))$
\IF{no improvement in the reconstruction loss in $p$ consecutive iterations}
\STATE early stop and exit 
\ENDIF
\STATE $k = k+1$
\ENDWHILE
\ENSURE the reconstructed image $x^{(k-p+1)}$ 
\end{algorithmic}
\end{algorithm}

\textbf{Our method: ES by self-validation}  \quad 
We now only need to detect the valley of the AE loss curve due to the blessed alignment discussed above. To sum up our algorithm: we train the structural OPT and an AE side-by-side, feed the most recent $n$ reconstructed images to train the AE, and then test the next image for AE loss. We make two additional modifications for better efficiency: 1) the AE is updated only once per new training set instead of being greedily trained, i.e., an online learning setting. This helps to substantially cut down the learning cost and save hyperparameters while remaining effective; and 2) to ensure the AEs learn the most compact representations, we borrow the idea of IRMAE~\cite{jing2020implicit} and add several trainable linear layers in the AEs before feeding the hidden code to the decoder. Our whole algorithmic pipeline is summarized in~\cref{alg:framework}.

\section{Experiments}
\label{sec:exps}

\textbf{Setup}  \quad 
We experiment with $3$ SIDGPs (DIP~\footnote{\url{https://github.com/DmitryUlyanov/deep-image-prior}}, DD\footnote{\url{https://github.com/reinhardh/supplement_deep_decoder}}, SIREN\footnote{\url{https://vsitzmann.github.io/siren/}}) and $4$ IR problems (image denoising, inpainting, MRI reconstruction, and image regression). Unless stated otherwise, we work with their default DNN architectures, optimizers, and hyperparameters. We fix the window size $n=256$, and the patience number $p = 500$ for denoising and inpainting and $p = 200$ for MRI reconstruction and image regression. For training the collaborative AEs, ADAM is used with a learning rate $10^{-3}$. To assess the IR quality, we use both PSNR~\cite{chan2005salt} and SSIM~\cite{wang2004image}. To evaluate our detection performance, we use ES-PG which is the absolute difference between the detected and the true peak PSNRs; similarly for ES-SG. Sometimes we also report BASELINE-PG which measures the difference between the peak and the final overfitting PSNRs; similarly for BASELINE-SG. For most small-scale experiments, we repeat all experiments $3$ times and report means and standard deviations. \textbf{All omitted details and experimental results can be found in the Appendix}. 


\begin{figure}[!htbp]
\vspace{-0.5em}
\begin{center}
{\includegraphics[width=1.0\textwidth]{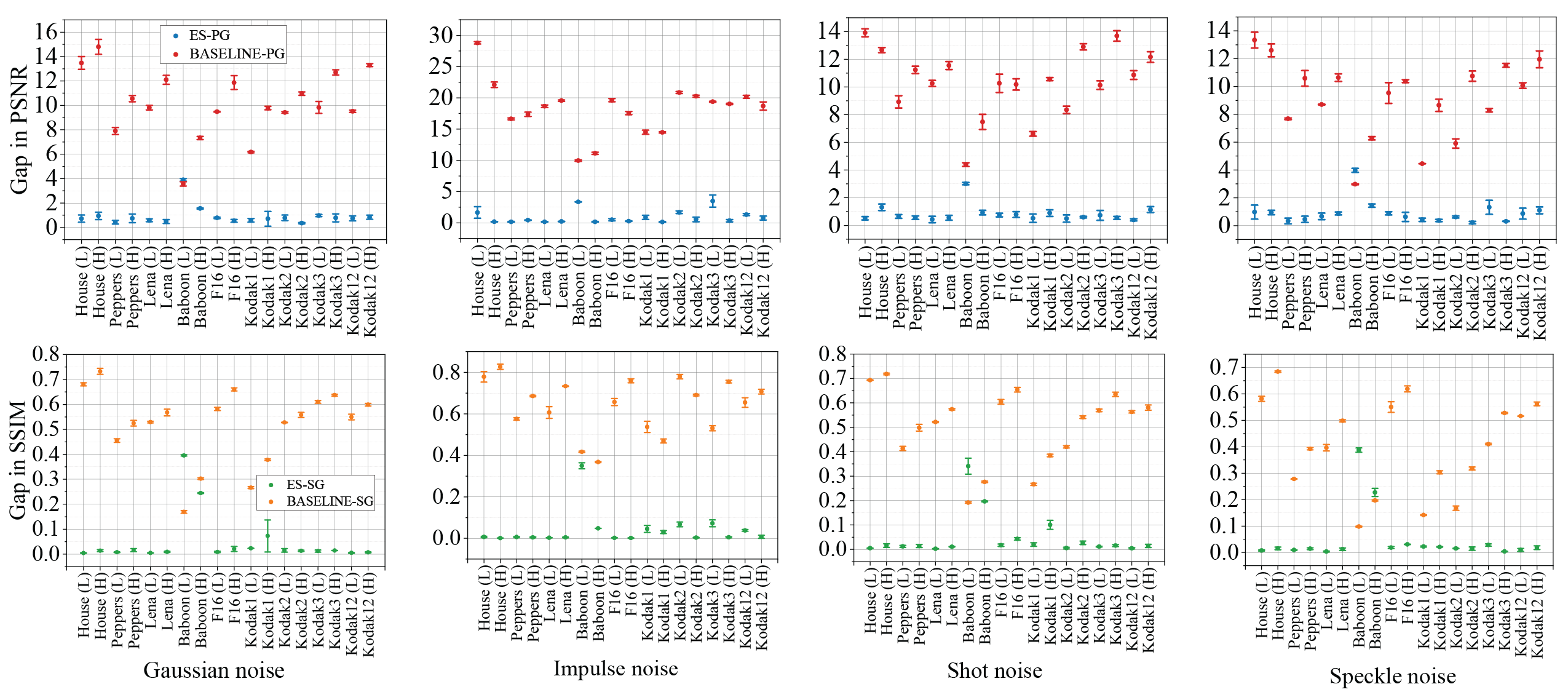}} 
\vspace{-3em}
\end{center}
   \caption{DIP+AE for image denoising. Each columns represents a distinct noise type, and the 1st row contains the PGs, and 2nd in SGs. The \emph{L} and \emph{H} in the horizontal annotations indicate low and high noise levels, respectively.}
   \vspace{-1em}
\label{fig:dip_denoising}
\end{figure}

\textbf{Image denoising}  \quad \label{subsec:image_denoising}
The power of DIP and DD was initially only demonstrated on Gaussian denoising. Here, to make the evaluation more thorough, we also experiment with denoising impulse, shot, and speckle noise, on a standard image denoising dataset\footnote{\url{http://www.cs.tut.fi/\~foi/GCF-BM3D/index.html\#ref_results}} ($9$ images). For each of the $4$ noise types, we test a low and a high noise level, respectively. The mean squares error (MSE) is used for Gaussian, shot, and speckle noise, while the $\ell_1$ loss is used for impulse noise. To obtain the final degraded results, we run DIP for $150$K iterations. 

The denoising results are measured in terms of the gap metrics are summarized in~\cref{fig:dip_denoising}. Note that our typical detection gap is $\le 1$ measured in ES-PG, and $\le 0.1$ measured in ES-SG. If DIP just runs without ES, the degradation of quality is severe, as indicated by both BASELINE-PG and BASELINE-SG. Evidently, our DIP+AE can save the computation and the reconstruction quality, and return an estimate with near-peak performance for almost all images, noise types, and noise levels that we test. The Baboon image is an outlier. It contains substantial high-frequency textures, and actually the peak PSNR/SSIM itself is also much worse compared to other images. We suspect that dealing with similar kinds of images can be challenging for DIP and DD and our detection method, which warrants future study. 

\begin{figure}[!htbp]
\vspace{-1em}
\begin{center}
{\includegraphics[width=1.0\textwidth]{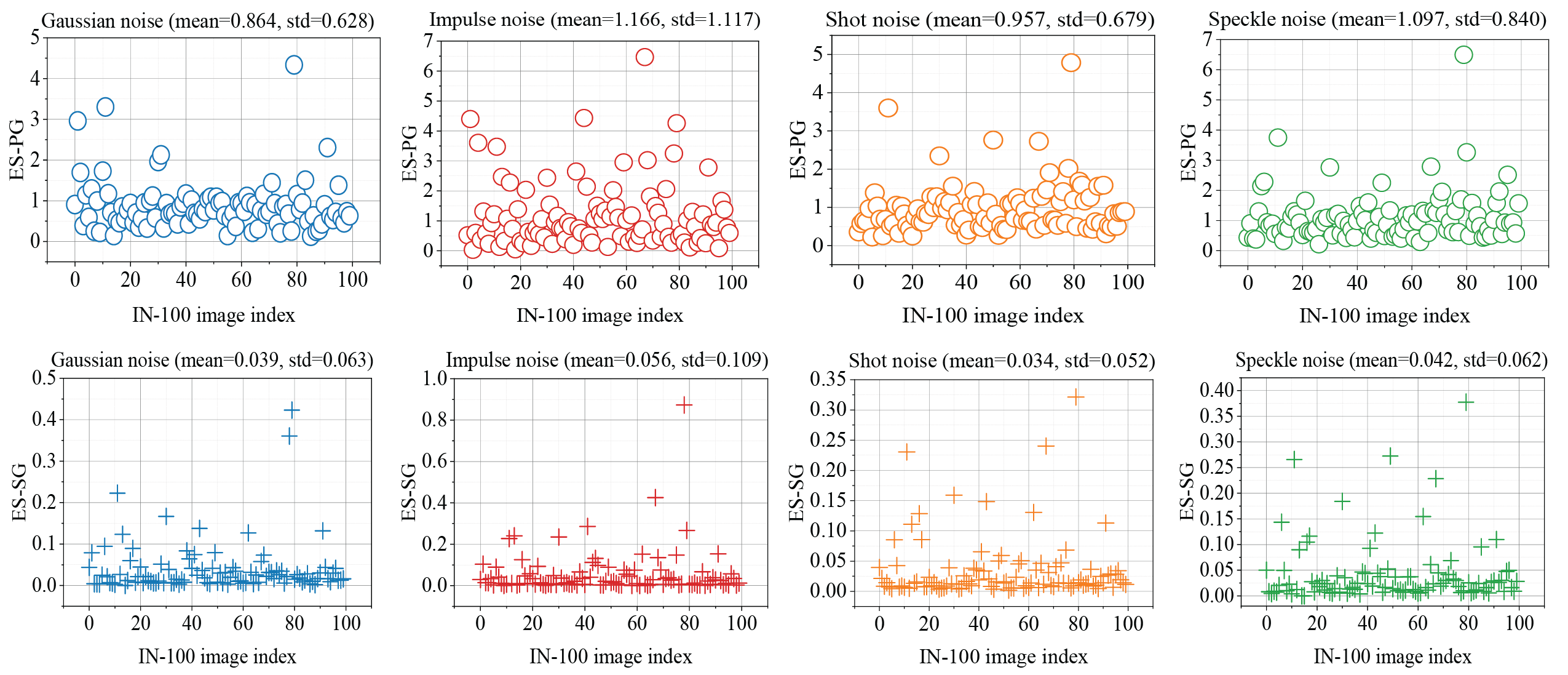}} 
\end{center}
 \vspace{-15pt}
\caption{DIP+AE on IN-100. 1st row: ES-PGs; 2nd row: ES-SGs.}
\label{fig:online_imagenet}
\end{figure}
We further test our method on $100$ randomly selected images from ImageNet~\cite{ILSVRC15}, denoted as IN-100. We follow the same evaluation protocol as above, except that we only experiment a medium noise level and we do not estimate the means and standard deviations; the results are reported in~\cref{fig:online_imagenet}. It is easy to see that the ES-PGs are concentrated around $1$ and the ES-GSs are concentrated around $0.1$, consistent with our observation on the small-scale dataset above. 

\begin{table*}[t]
\caption{DIP+AE, DIP+TV, and SGLD on Gaussian denoising. The best PSNRs are colored as \textcolor{red}{red}; the best SSIMs are colored as \textcolor{blue}{blue}.}
\vspace{-0.5em}
\label{tab:denoising_comp}
\begin{center}
\setlength{\tabcolsep}{0.5mm}{
\begin{tabular}{l c c c| c c c}
\hline
\multirow{2}{*}{\textit{}} & 
\multicolumn{3}{c|}{\scriptsize{PSNR $\uparrow$}} &
\multicolumn{3}{c}{\scriptsize{SSIM $\uparrow$}}
\\

\hline
&
\scriptsize{DIP+AE} &
\scriptsize{DIP+TV} &
\scriptsize{SGLD} &
\scriptsize{DIP+AE} &
\scriptsize{DIP+TV} &
\scriptsize{SGLD} 
 \\

\hline
\scriptsize{House}
&\tabincell{c}{\textcolor{red}{\scriptsize{30.824}}\tiny{(0.236)}}
&\tabincell{c}{\scriptsize{18.866}\tiny{(0.093)}}
&\tabincell{c}{\scriptsize{21.800}\tiny{(2.430)}}
&\tabincell{c}{\textcolor{blue}{\scriptsize{0.834}} \tiny{(0.008)}}
&\tabincell{c}{\scriptsize{0.174} \tiny{(0.006)}}
&\tabincell{c}{\scriptsize{0.495} \tiny{(0.152)}}
\\

\hline
\scriptsize{Peppers}
&\tabincell{c}{\textcolor{red}{\scriptsize{26.471}} \tiny{(0.249)}}
&\tabincell{c}{\scriptsize{19.000} \tiny{(0.004)}}
&\tabincell{c}{\scriptsize{20.516} \tiny{(0.263)}}
&\tabincell{c}{\textcolor{blue}{\scriptsize{0.688}} \tiny{(0.008)}}
&\tabincell{c}{\scriptsize{0.238} \tiny{(0.001)}}
&\tabincell{c}{\scriptsize{0.453} \tiny{(0.017)}}
\\

\hline
\scriptsize{Lena}
&\tabincell{c}{\textcolor{red}{\scriptsize{27.462}} \tiny{(0.602)}}
&\tabincell{c}{\scriptsize{18.961} \tiny{(0.028)}}
&\tabincell{c}{\scriptsize{20.217} \tiny{(0.808)}}
&\tabincell{c}{\textcolor{blue}{\scriptsize{0.744}} \tiny{(0.003)}}
&\tabincell{c}{\scriptsize{0.253} \tiny{(0.001)}}
&\tabincell{c}{\scriptsize{0.447} \tiny{(0.051)}}
\\

\hline
\scriptsize{Baboon}
&\tabincell{c}{\textcolor{red}{\scriptsize{19.453}} \tiny{(0.251)}}
&\tabincell{c}{\scriptsize{18.416} \tiny{(0.024)}}
&\tabincell{c}{\scriptsize{19.111} \tiny{(0.091)}}
&\tabincell{c}{\textcolor{blue}{\scriptsize{0.335}} \tiny{(0.008)}}
&\tabincell{c}{\scriptsize{0.467} \tiny{(0.0)}}
&\tabincell{c}{\scriptsize{0.572} \tiny{(0.003)}}
\\

\hline
\scriptsize{F16}
&\tabincell{c}{\textcolor{red}{\scriptsize{27.644}} \tiny{(0.114)}}
&\tabincell{c}{\scriptsize{19.183} \tiny{(0.122)}}
&\tabincell{c}{\scriptsize{20.431} \tiny{(0.091)}}
&\tabincell{c}{\textcolor{blue}{\scriptsize{0.818}} \tiny{(0.005)}}
&\tabincell{c}{\scriptsize{0.248} \tiny{(0.005)}}
&\tabincell{c}{\scriptsize{0.457} \tiny{(0.008)}}
\\

\hline
\scriptsize{Kodak1}
&\tabincell{c}{\textcolor{red}{\scriptsize{24.455}} \tiny{(0.111)}}
&\tabincell{c}{\scriptsize{18.688} \tiny{(0.088)}}
&\tabincell{c}{\scriptsize{19.555} \tiny{(0.049)}}
&\tabincell{c}{\textcolor{blue}{\scriptsize{0.649}} \tiny{(0.006)}}
&\tabincell{c}{\scriptsize{0.409} \tiny{(0.006)}}
&\tabincell{c}{\scriptsize{0.528} \tiny{(0.003)}}
\\

\hline
\scriptsize{Kodak2}
&\tabincell{c}{\textcolor{red}{\scriptsize{26.886}} \tiny{(0.218)}}
&\tabincell{c}{\scriptsize{19.535} \tiny{(0.160)}}
&\tabincell{c}{\scriptsize{20.569} \tiny{(0.629)}}
&\tabincell{c}{\textcolor{blue}{\scriptsize{0.677}} \tiny{(0.006)}}
&\tabincell{c}{\scriptsize{0.243} \tiny{(0.014)}}
&\tabincell{c}{\scriptsize{0.430} \tiny{(0.046)}}
\\

\hline
\scriptsize{Kodak3}
&\tabincell{c}{\textcolor{red}{\scriptsize{27.894}} \tiny{(0.311)}}
&\tabincell{c}{\scriptsize{18.904} \tiny{(0.181)}}
&\tabincell{c}{\scriptsize{19.921} \tiny{(0.194)}}
&\tabincell{c}{\textcolor{blue}{\scriptsize{0.761}} \tiny{(0.002)}}
&\tabincell{c}{\scriptsize{0.187} \tiny{(0.008)}}
&\tabincell{c}{\scriptsize{0.379} \tiny{(0.012)}}
\\

\hline
\scriptsize{Kodak12}
&\tabincell{c}{\textcolor{red}{\scriptsize{28.269}} \tiny{(0.239)}}
&\tabincell{c}{\scriptsize{18.998} \tiny{(0.183)}}
&\tabincell{c}{\scriptsize{19.974} \tiny{(0.240)}}
&\tabincell{c}{\textcolor{blue}{\scriptsize{0.727}} \tiny{(0.003)}}
&\tabincell{c}{\scriptsize{0.193} \tiny{(0.011)}}
&\tabincell{c}{\scriptsize{0.391} \tiny{(0.015)}}
\\
\hline
\end{tabular}
}
\end{center}
\vspace{-2em}
\end{table*}

We compare DIP+AE with three other competing methods that try to eliminate overfitting, i.e., DIP+TV~\cite{sun2020solving, sun2021plug} and SGLD~\cite{welling2011bayesian} on Gaussian denoising, and DOP~\cite{NEURIPS2020_cd42c963} on removing sparse corruptions. We take a medium noise level. All these methods claim to eliminate the overfitting altogether, and so we directly compare the detected PSNRs (SSIMs) by our method and the final PSNRs returned by their methods at the final convergence. The results are tabulated in \cref{tab:denoising_comp} and \cref{tab:dop_vs_dop}, respectively. It seems that 1) DIP+TV and SGLD do not quite eliminate overfitting when different images and noise levels (noise levels used in the respective papers are relatively low) than they experimented with are used; and 2) our detection method, i.e., DIP+AE, returns far better reconstruction than DIP+TV and SGLD (except on Baboon), and comparable results to DOP on sparse corruptions (in our experiments, we do not add any perturbations to the input random noise of DOP). 
\begin{table*}[!htpb]
\begin{center}
\caption{DOP vs. DIP+AE on removing sparse corruptions}
\label{tab:dop_vs_dop}
\setlength{\tabcolsep}{0.15mm}{
\begin{tabular}{c c| c c c c c c c c c}
\hline
\scriptsize{Metrics} &\scriptsize{Models} &\scriptsize{House} &\scriptsize{Peppers} &\scriptsize{Lena} &\scriptsize{Baboon} &\scriptsize{F16} &\scriptsize{Kodak1} &\scriptsize{Kodak2} &\scriptsize{Kodak3} &\scriptsize{Kodak12}
\\
\hline
\multirow{2}{*}{\scriptsize{PSNR $\uparrow$}} 

& \scriptsize{DOP}
& \scriptsize{35.148}
& \scriptsize{{28.635}}
& \scriptsize{30.592}
& \scriptsize{{21.051}}
& \scriptsize{{28.918}}
& \scriptsize{{25.006}}
& \scriptsize{29.516}
& \scriptsize{30.098}
& \scriptsize{29.694}\\

& \scriptsize{DIP+AE}
& \scriptsize{{38.343}} 
& \scriptsize{27.724}
& \scriptsize{{30.768}}
& \scriptsize{19.476}
& \scriptsize{28.253}
& \scriptsize{24.954}
& \scriptsize{29.812}
& \scriptsize{30.220}
& \scriptsize{30.883}\\

\hline
\multirow{2}{*}{\scriptsize{SSIM $\uparrow$}} 

& \scriptsize{DOP}
& \scriptsize{0.914}
& \scriptsize{0.753}
& \scriptsize{0.831}
& \scriptsize{0.578}
& \scriptsize{0.869}
& \scriptsize{0.690}
& \scriptsize{0.766}
& \scriptsize{0.843}
& \scriptsize{0.803}\\

& \scriptsize{DIP+AE}
& \scriptsize{0.954}
& \scriptsize{0.783}
& \scriptsize{0.845}
& \scriptsize{0.377}
& \scriptsize{0.892}
& \scriptsize{0.723}
& \scriptsize{0.785}
& \scriptsize{0.861}
& \scriptsize{0.844}\\

\hline
\multirow{2}{*}{\scriptsize{Iteration $\downarrow$} }
& \scriptsize{DOP}
& \scriptsize{149740}
& \scriptsize{149469}
& \scriptsize{141000}
& \scriptsize{37000}
& \scriptsize{149370}
& \scriptsize{147096}
& \scriptsize{145595}
& \scriptsize{148666}
& \scriptsize{141390}\\

& \scriptsize{DIP+AE}
& \scriptsize{3425}
& \scriptsize{1405}
& \scriptsize{1638}
& \scriptsize{360}
& \scriptsize{1524}
& \scriptsize{1322}
& \scriptsize{1038}
& \scriptsize{1112}
& \scriptsize{1522}\\
\hline
\end{tabular}
}
\end{center}
\vspace{-1.5em}
\end{table*}

\begin{figure*}[!htbp]
\begin{center}
{\includegraphics[width=0.8\textwidth]{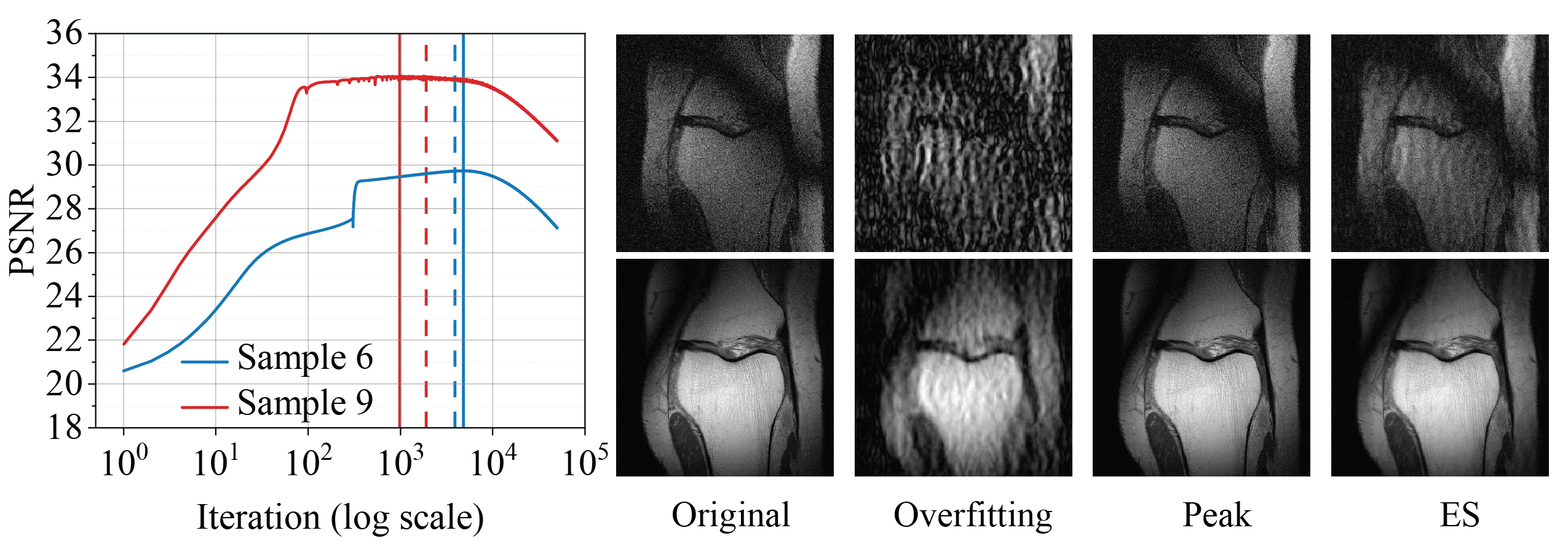}} 
\end{center}
\vspace{-15pt}
\caption{Results for MRI reconstruction. (left) The solid vertical lines indicate the peak performance iterate while the dash vertical lines are ES iterate detected by our method. (right) Visualizations for Sample 6 (1st row) and Sample 9 (2nd row).}
\label{fig:mri_curve_image}
\end{figure*}

We further compare our method with a group of baseline methods based on no-reference image quality metrics: the classical BRISQUE~\cite{mittal2012no} and NIQE~\cite{mittal2012making}, and a state-of-the-art, NIMA (both technical quality and aesthetic assessment)~\cite{talebi2018nima} which is based on pretrained DNNs. We experiment with medium noise level across all four noise types with DIP, run DIP until the final convergence (i.e., overfitting), and then select the highest quality one from among all the intermediate reconstructions according to each of the three metrics, respectively. Performance gaps on medium Gaussian and impluse noise are presented in~\cref{tab:other_metric_part1}, on shot noise and speckle noise presented in~\cref{tab:other_metric_part2} in the Appendix. Told from \cref{tab:other_metric_part1} and~\cref{tab:other_metric_part2}, our method is a clear winner: in most cases, our method has the smallest gaps. Moreover, while our detection gaps are uniformly small (with rare outliers), the three baseline methods can frequently lead to intolerably large gaps. 
\begin{table*}[!htpb]
\caption{The performance gaps of BRISQUE~\cite{mittal2012no}, NIQE~\cite{mittal2012making}, NIMA~\cite{talebi2018nima}, and DIP+AE on Gaussian and impulse noises. For NIMA, we report both technical quality assessment (the number before ``/'') and aesthetic assessment (the number after ``/''). The best PSNR gaps are colored as \textcolor{red}{red}; the best SSIM gaps are colored as \textcolor{blue}{blue}.}
\vspace{-1em}
\label{tab:other_metric_part1}
\begin{center}
\setlength{\tabcolsep}{0.35mm}{
\begin{tabular}{l c c c c| c c c c|| c c c c| c c c c}
& 
\multicolumn{8}{c}{\scriptsize{Gaussian noise}} &
\multicolumn{8}{c}{\scriptsize{Impulse noise}}
\\
\hline
&\multicolumn{4}{c|}{\tiny{Gap in PSNR $\downarrow$}} 
&\multicolumn{4}{c||}{\tiny{Gap in SSIM $\downarrow$}} 
&\multicolumn{4}{c|}{\tiny{Gap in PSNR $\downarrow$}} 
&\multicolumn{4}{c}{\tiny{Gap in SSIM $\downarrow$}}
\\
\hline
& \tiny{BRISQUE}
& \tiny{NIQE}
& \tiny{NIMA}
& \tiny{DIP+AE}

& \tiny{BRISQUE}
& \tiny{NIQE}
& \tiny{NIMA}
& \tiny{DIP+AE}

& \tiny{BRISQUE}
& \tiny{NIQE}
& \tiny{NIMA}
& \tiny{DIP+AE}

& \tiny{BRISQUE}
& \tiny{NIQE}
& \tiny{NIMA}
& \tiny{DIP+AE}
\\
\hline
\tiny{House}
&\tiny{10.961}
&\tiny{10.961}
&\tiny{12.906/3.361}
&\tiny{\textcolor{red}{1.057}}

&\tiny{0.408}
&\tiny{0.408}
&\tiny{0.659/0.122}
&\tiny{\textcolor{blue}{0.015}}

&\tiny{25.227}
&\tiny{9.886}
&\tiny{25.227/21.403}
&\tiny{\textcolor{red}{5.832}}

&\tiny{0.835}
&\tiny{0.052}
&\tiny{0.835/0.344}
&\tiny{\textcolor{blue}{0.005}}
\\
\hline
\tiny{Peppers}
&\tiny{3.715}
&\tiny{5.053}
&\tiny{4.893/5.204}
&\tiny{\textcolor{red}{0.603}}

&\tiny{0.271}
&\tiny{0.341}
&\tiny{0.334/0.345}
&\tiny{\textcolor{blue}{0.006}}

&\tiny{12.872}
&\tiny{2.296}
&\tiny{2.117/9.701}
&\tiny{\textcolor{red}{0.305}}

&\tiny{0.404}
&\tiny{0.013}
&\tiny{\textcolor{blue}{0.012}/0.198}
&\tiny{\textcolor{blue}{0.012}}
\\

\hline
\tiny{Lena}
&\tiny{4.681}
&\tiny{7.194}
&\tiny{10.180/1.581}
&\tiny{\textcolor{red}{0.845}}

&\tiny{0.260}
&\tiny{0.397}
&\tiny{0.537/0.054}
&\tiny{\textcolor{blue}{0.008}}

&\tiny{2.374}
&\tiny{6.204}
&\tiny{15.646/12.674}
&\tiny{\textcolor{red}{0.231}}

&\tiny{0.011}
&\tiny{0.039}
&\tiny{0.376/0.209}
&\tiny{\textcolor{blue}{0.002}}
\\

\hline
\tiny{Baboon}
&\tiny{\textcolor{red}{0.247}}
&\tiny{0.694}
&\tiny{1.375/3.450}
&\tiny{2.583}

&\tiny{\textcolor{blue}{0.008}}
&\tiny{0.014}
&\tiny{0.174/0.416}
&\tiny{0.304}

&\tiny{7.092}
&\tiny{\textcolor{red}{1.312}}
&\tiny{1.157/12.707}
&\tiny{1.426}

&\tiny{0.224}
&\tiny{\textcolor{blue}{0.013}}
&\tiny{0.153/0.462}
&\tiny{0.231}
\\

\hline
\tiny{F16}
&\tiny{6.236}
&\tiny{8.128}
&\tiny{1.336/3.991}
&\tiny{\textcolor{red}{0.850}}

&\tiny{0.429}
&\tiny{0.530}
&\tiny{0.036/0.088}
&\tiny{\textcolor{blue}{0.023}}

&\tiny{14.180}
&\tiny{5.505}
&\tiny{8.621/9.372}
&\tiny{\textcolor{red}{0.495}}

&\tiny{0.369}
&\tiny{0.036}
&\tiny{0.111/0.135}
&\tiny{\textcolor{blue}{0.006}}
\\

\hline
\tiny{Kodak1}
&\tiny{2.558}
&\tiny{3.158}
&\tiny{6.723/19.546}
&\tiny{\textcolor{red}{0.520}}

&\tiny{0.069}
&\tiny{0.103}
&\tiny{0.275/0.626}
&\tiny{\textcolor{blue}{0.029}}

&\tiny{5.913}
&\tiny{\textcolor{red}{0.232}}
&\tiny{11.791/10.366}
&\tiny{0.538}

&\tiny{0.142}
&\tiny{\textcolor{blue}{0.007}}
&\tiny{0.410/0.584}
&\tiny{0.048}
\\

\hline
\tiny{Kodak2}
&\tiny{6.959}
&\tiny{6.855}
&\tiny{9.094/\textcolor{red}{0.143}}
&\tiny{0.353}

&\tiny{0.361}
&\tiny{0.365}
&\tiny{0.516/\textcolor{blue}{0.004}}
&\tiny{0.007}

&\tiny{18.575}
&\tiny{\textcolor{red}{0.312}}
&\tiny{13.870/9.657}
&\tiny{0.429}

&\tiny{0.741}
&\tiny{\textcolor{blue}{0.014}}
&\tiny{0.304/0.127}
&\tiny{0.026}
\\

\hline
\tiny{Kodak3}
&\tiny{2.029}
&\tiny{7.686}
&\tiny{19.834/19.716}
&\tiny{\textcolor{red}{0.574}}

&\tiny{0.152}
&\tiny{0.475}
&\tiny{0.731/0.624}
&\tiny{\textcolor{blue}{0.005}}

&\tiny{16.867}
&\tiny{8.769}
&\tiny{13.217/2.031}
&\tiny{\textcolor{red}{1.959}}

&\tiny{0.351}
&\tiny{0.077}
&\tiny{0.180/\textcolor{blue}{0.005}}
&\tiny{0.046}
\\

\hline
\tiny{Kodak12}
&\tiny{7.066}
&\tiny{8.093}
&\tiny{6.436/3.075}
&\tiny{\textcolor{red}{0.788}}

&\tiny{0.419}
&\tiny{0.456}
&\tiny{0.384/0.178}
&\tiny{\textcolor{blue}{0.007}}

&\tiny{16.781}
&\tiny{1.785}
&\tiny{9.139/2.685}
&\tiny{\textcolor{red}{1.068}}

&\tiny{0.817}
&\tiny{\textcolor{blue}{0.003}}
&\tiny{0.087/0.008}
&\tiny{0.048}
\\
\hline
\end{tabular}
}
\end{center}
\end{table*}

\textbf{MRI reconstruction} \quad \label{subsec:mri}
We now test our detection method on MRI reconstruction, a classical medical IR problem involving a nontrivial linear $f$. Specifically, the model is $y = f(x)+\xi= \mc F \paren{x} + \xi$, where $\mc F$ is the subsampled Fourier operator and $\xi$ models the noise encountered in practical MRI imaging. Here, we take $8$-fold undersampling and choose to parametrize $x$ using a DD, inspired by the recent work~\cite{pmlr-v119-heckel20a}. Due to the heavy overparametrization of DD, \cite{pmlr-v119-heckel20a} needs to choose an appropriate ES to produce quality reconstruction. We report the performance in~\cref{fig:mri_curve_image} (results for all randomly selected samples can be found in the Appendix). Our method is able to signal stopping points that are reasonably close to the peak points, which also yield reasonably faithful reconstruction.


\begin{figure}
  \centering
  \includegraphics[width=0.8\linewidth]{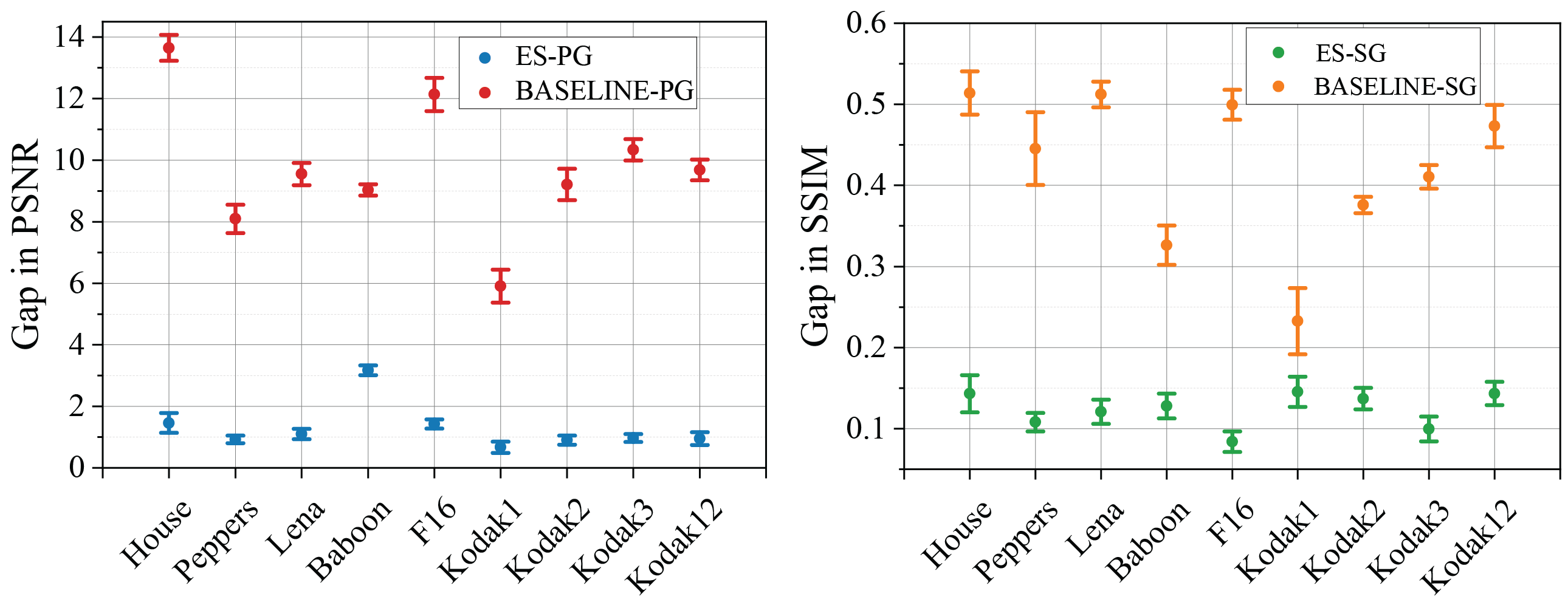}
  \caption{Results for image regression.}
  \label{fig:image_regression}
\end{figure}

\textbf{Image regression} \quad 
\label{subsec:img_reg}
Now we turn to SIREN~\cite{sitzmann2020implicit}, a recent functional SIDGP model that is designed to facilitate the learning of functions with significant high-frequency components. We consider a simple task from the original task, image regression, but add in some Gaussian noise. Mathematically, the $y = x + \eps$, where $\eps \sim_{iid} \mc N\paren{0, 0.196}$, and we are to fit $y$ by performing functional OPT listed in \cref{eq:inv_obj_2}. Clearly, when the MLP used in SIREN is sufficiently overparamterized, the noise will also be learned. We test our detection method on this using the same $9$-image dataset as in denoising. From \cref{fig:image_regression}, we can see again that our method is capable of reliably detecting near-peak performance measured by either ES-PG or ES-SG, much better than without implementing any ES.

\textbf{Ablation studies} \quad \label{subsec:ablation studies}
\begin{figure*}[!htbp]
\begin{center}
{\includegraphics[width=0.75\textwidth]{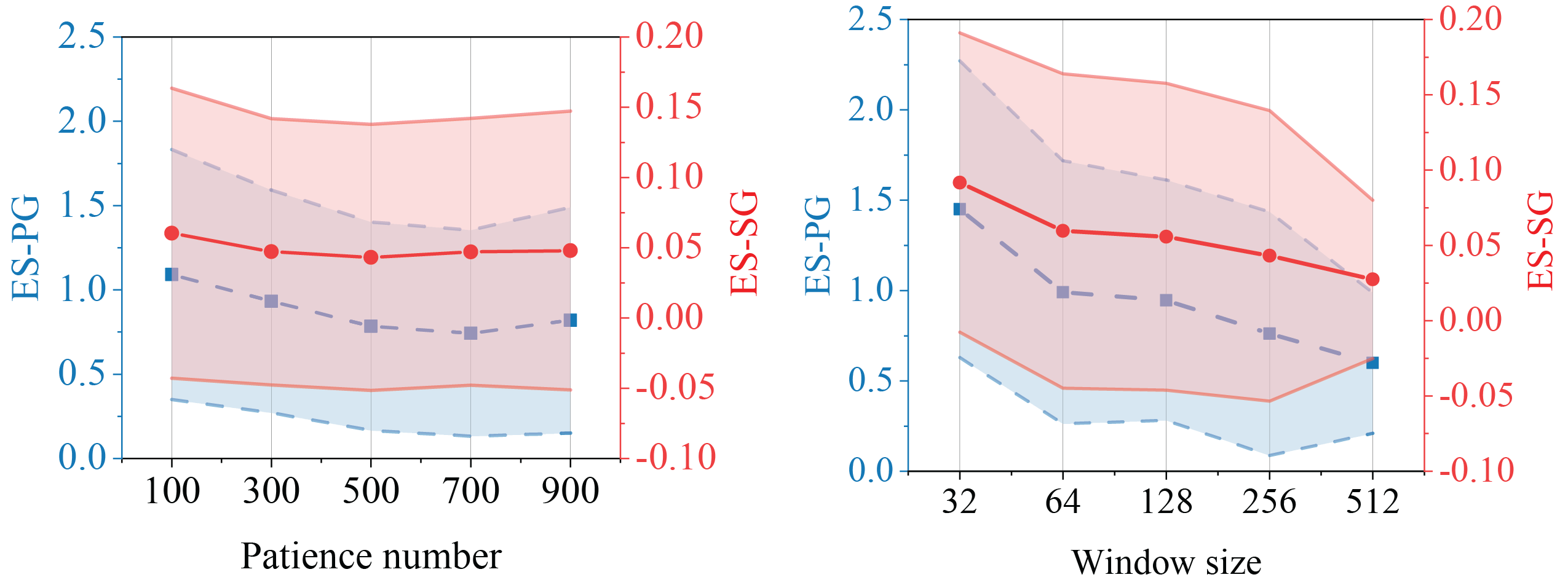}} 
\vspace{-18pt}
\end{center}
\caption{Ablation study for the sensitivity of the detection performance with respect to the patience number $p$ and window size $n$. }
\label{fig:pnum_vs_ws}
\vspace{-0.5em}
\end{figure*}
There are two crucial parameters---the window size $n$ and the patience number $p$---dictating the performance of our detection method. Here, we evaluate the sensitivity of the performance to $n$ and $p$ on two settings: 1) fix $n=256$, and vary $p$ in $\{100, 300, 500, 700, 900\}$; 2) fix $p=500$, and vary $n$ in $\{32, 64, 128, 256, 512\}$. We again use the $9$-image dataset, simulate medium-level Gaussian noise, and adopt DIP. For simplicity, we only report the gaps averaged over all images. \cref{fig:pnum_vs_ws} summarizes the results. Two observations: 1) our detection method is relatively insensitive to the two parameters; 2) large $p$ and $n$ seem to benefit the performance more. To be sure, we need $p$ to be at least reasonably large so that our detection will not be trapped by local spikes. However, larger $n$ requires better GPU resources to hold and compute with the data. In addition, we also explore the influence of the learing rates---for SIDGP and AE, respectively---on the performance. For this, we again consider medium Gaussian noise with DIP, and freeze $n = 256$ and $p= 500$. We fix one of the two learning rates while changing the other, and present the results in \cref{tab:diff_LR}. 
We observe that despite different learning rates lead to slightly different levels of performance, the variation is not significant, in view that our typical ES-PG is less than $1$, and typical ES-SG less than $0.1$ in our previous evaluation. 
\begin{table*}[!htpb]
\caption{The impacts of different learning rates. }
\vspace{-0.5em}
\label{tab:diff_LR}
\begin{center}
\setlength{\tabcolsep}{5mm}{
\begin{tabular}{c| c c| c c}
\hline
\multirow{2}{*}{\scriptsize{Learning Rate}}
&
\multicolumn{2}{c|}{\scriptsize{(vary) DIP + (fixed $.001$) AE}} &
\multicolumn{2}{c}{\scriptsize{(fixed $.01$) DIP + (vary) AE}}
\\

&\scriptsize{ES-PG $\downarrow$} 
&\scriptsize{ES-SG $\downarrow$} 
&\scriptsize{ES-PG $\downarrow$} 
&\scriptsize{ES-SG $\downarrow$}
\\
\hline
\scriptsize{0.01}
& \scriptsize{0.756} (\tiny{0.629})
& \scriptsize{0.044} (\tiny{0.093})
& \scriptsize{0.814} (\tiny{0.623})
& \scriptsize{0.043} (\tiny{0.090})
\\

\hline
\scriptsize{0.001}
& \scriptsize{0.593} (\tiny{0.497})
& \scriptsize{0.043} (\tiny{0.085})
& \scriptsize{0.713} (\tiny{0.628})
& \scriptsize{0.045} (\tiny{0.098})
\\

\hline
\scriptsize{0.0001}
& \scriptsize{1.073} (\tiny{0.984})
& \scriptsize{0.074} (\tiny{0.118})
& \scriptsize{1.216} (\tiny{1.036})
& \scriptsize{0.071} (\tiny{0.098})
\\

\hline
\end{tabular}
}
\end{center}
\end{table*}

\newpage 

\section*{Acknowledgement}
Zhong Zhuang, Hengkang Wang, and Ju Sun are partly supported by NSF CMMI 2038403. The authors acknowledge the Minnesota Supercomputing Institute (MSI) at the University of Minnesota for providing resources that contributed to the research results reported within this paper.

\bibliography{references,ref}

\appendix
\newpage
\section{Additional experimental details}\label{appendix:framework_noise}
In this section, we provide experimental details omitted in the main body of the paper. 

\begin{itemize}
    
    \item \emph{Noise settings}. We simulate $4$ different types of noise and $3$ intensity levels for each noise type, as detailed below. The generation algorithms follow those of ImageNet-C~\cite{hendrycks2018benchmarking} \footnote{\url{https://github.com/hendrycks/robustness}}. 
    \begin{itemize}
    	\item Gaussian noise: $0$ mean additive Gaussian noise with variance $0.12$, $0.18$, and $0.26$ for low, medium, and high noise levels, respectively; 
    	\item Impulse noise: i.e., salt-and-pepper noise, replacing each pixel with probability $p \in [0, 1]$ into white or black pixel with half chance each. Low, medium, and high noise levels correspond to $p = 0.3, 0.5, 0.7$, respectively; 
    	\item Shot noise: i.e., pixel-wise independent Poisson noise. For each pixel $x \in [0, 1]$, the noisy pixel is Poisson distributed with rate $\lambda x$, where $\lambda$ is $25, 12, 5$ for low, medium, and high noise levels, respectively.  
    	\item Speckle noise: for each pixel $x \in [0, 1]$, the noisy pixel is $x\paren{1+\eps}$, where $\eps$ is 0-mean Gaussian with a variance level $0.20$, $0.35$, $0.45$ for low, medium, and high noise levels, respectively. 
    \end{itemize}
    
    \item \emph{Network architecture}. Our AE is based on deep CNNs. The exact architecture is depicted in~\cref{tab:AE_net}.
    
\end{itemize}



\begin{table*}[!htpb]
\caption{The network architecture of AE.}
\begin{center}
\begin{threeparttable}
\setlength{\tabcolsep}{8mm}{
\begin{tabular}{c c c}
\hline
\scriptsize{Nets} 
&\scriptsize{Layers}
&\scriptsize{Parameters}\\

\hline
\multirow{14}{*}{ \rotatebox{90}{\scriptsize{Encoder net}}} 
& \scriptsize{Conv2d~\tnote{1}} & \scriptsize{(3, 32, 3, 2, 1, False)}\\
& \scriptsize{Batch norm, ReLU} & \scriptsize{N/A}\\

& \scriptsize{Conv2d} & \scriptsize{(32, 64, 3, 2, 1, False)}\\
& \scriptsize{Batch norm, ReLU} & \scriptsize{N/A}\\

& \scriptsize{Conv2d} & \scriptsize{(64, 128, 3, 2, 1, False)}\\
& \scriptsize{Batch norm, ReLU} & \scriptsize{N/A}\\

& \scriptsize{Conv2d} & \scriptsize{(128, 128, 3, 2, 1, False)}\\
& \scriptsize{Batch norm, ReLU} & \scriptsize{N/A}\\

& \scriptsize{Conv2d} & \scriptsize{(128, 128, 3, 2, 1, False)}\\
& \scriptsize{Batch norm, ReLU} & \scriptsize{N/A}\\

& \scriptsize{Conv2d} & \scriptsize{(128, 128, 3, 2, 1, False)}\\
& \scriptsize{Batch norm, ReLU} & \scriptsize{N/A}\\

& \scriptsize{Conv2d} & \scriptsize{(128, 1, 3, 2, 1, False)}\\
& \scriptsize{Batch norm, ReLU} & \scriptsize{N/A}\\

\hline
\hline
\multirow{4}{*}{ \rotatebox{90}{{\scriptsize{Linear net~\tnote{3}}}}} 
& \scriptsize{Linear~\tnote{2}} & \scriptsize{(16, 16, False)}\\
& \scriptsize{Linear} & \scriptsize{(16, 16, False)}\\
& \scriptsize{Linear} & \scriptsize{(16, 16, False)}\\
& \scriptsize{Linear} & \scriptsize{(16, 16, False)}\\
\hline
\hline
\multirow{21}{*}{ \rotatebox{90}{\scriptsize{Decoder net~\tnote{3}}}} 
& \scriptsize{Upsample}  & \scriptsize{bilinear}\\
& \scriptsize{Conv2d} & \scriptsize{(1, 128, 3, 1, 1, False)}\\
& \scriptsize{Batch norm, ReLU} & \scriptsize{N/A}\\

& \scriptsize{Upsample}  & \scriptsize{bilinear}\\
& \scriptsize{Conv2d} & \scriptsize{(128, 128, 3, 1, 1, False)}\\
& \scriptsize{Batch norm, ReLU} & \scriptsize{N/A}\\

& \scriptsize{Upsample}  & \scriptsize{bilinear}\\
& \scriptsize{Conv2d} & \scriptsize{(128, 128, 3, 1, 1, False)}\\
& \scriptsize{Batch norm, ReLU} & \scriptsize{N/A}\\

& \scriptsize{Upsample}  & \scriptsize{bilinear}\\
& \scriptsize{Conv2d} & \scriptsize{(128, 128, 3, 1, 1, False)}\\
& \scriptsize{Batch norm, ReLU} & \scriptsize{N/A}\\

& \scriptsize{Upsample}  & \scriptsize{bilinear}\\
& \scriptsize{Conv2d} & \scriptsize{(128, 64, 3, 1, 1, False)}\\
& \scriptsize{Batch norm, ReLU} & \scriptsize{N/A}\\

& \scriptsize{Upsample}  & \scriptsize{bilinear}\\
& \scriptsize{Conv2d} & \scriptsize{(64, 32, 3, 1, 1, False)}\\
& \scriptsize{Batch norm, ReLU} & \scriptsize{N/A}\\

& \scriptsize{Upsample}  & \scriptsize{bilinear}\\
& \scriptsize{Conv2d} & \scriptsize{(32, 3, 3, 1, 1, False)}\\
& \scriptsize{Batch norm, Sigmoid} & \scriptsize{N/A}\\

\hline
\end{tabular}
}
\begin{tablenotes}
\footnotesize
\item[1] The parameters for Conv2d layers: (in\_channels, out\_channels, kernel\_size, stride, padding, bias).
\item[2] The parameters for Linear layers: (in\_features, out\_features, bias).
\item[3] Tensors are reshaped properly to suit the input dimensions.
\end{tablenotes}
\end{threeparttable}
\end{center}
\label{tab:AE_net}
\end{table*}



\section{Additional experimental results}

\subsection{Image denoising}\label{appendix:deep_decoder}

Besides DIP, we also verify our methods on DD. As we alluded to in \cref{sec:intro}, although the original DD paper proposes underparameterization as a strategy to tame overfitting, in practice it is tricky to implement and underparameterization can produce inferior results, see, e.g., \cref{fig:dip_dd_overfit_curve} (right). Thus, people (including the DD authors in their later papers, e.g., \cite{pmlr-v119-heckel20a}) tend to still use overparametrized DD. Empirically, overparametrized DDs behave very similarly to DIPs. Since in \cref{subsec:image_denoising} we have validated our detection method extensively on DIP with $4$ noise types of both low and high noise levels, here we only focus on Gaussian noise with medium noise level. Here, we set all the network width as $512$, which is typically used in practice. The learning rate is set to $0.001$ here for good numerical stability. Other experimental settings are identical to those of DIP in~\cref{sec:exps}. 

\begin{figure}[t]
  \centering
  \includegraphics[width=0.8\linewidth]{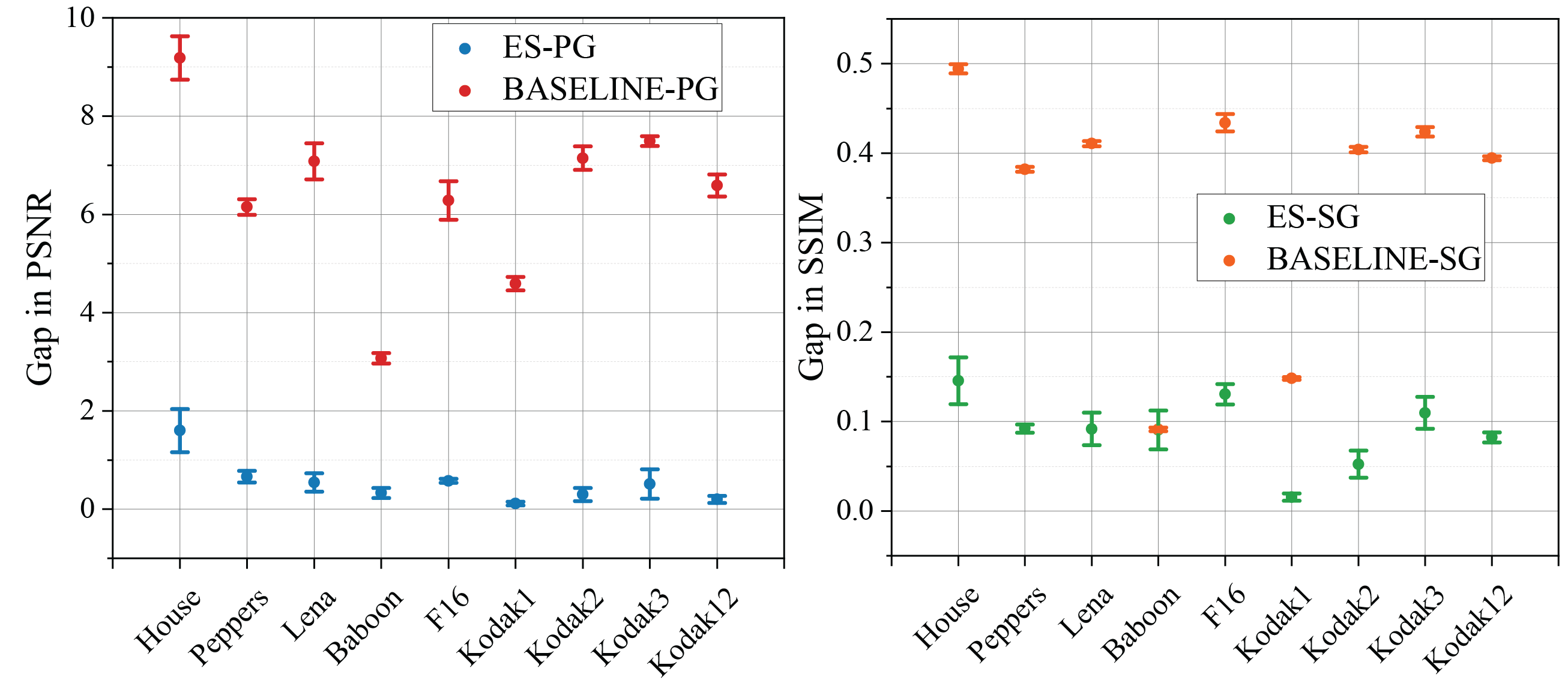}
  \caption{DD+AE for image denoising. (left) The performance measured in PGs. (right) The performance measured in SGs.}
  \label{fig:dd_denoising}
\end{figure}

The denoising results are summarized in~\cref{fig:dd_denoising}. One can observe that in most cases, the detection gap is $\le 1$ in terms of ES-PG, and $\le 0.1$ in terms of ES-SG. However, if we run DD without ES, the overfitting issue is dreadful: most of BASELINE-PGs are $\ge 6$ and BASELINE-SGs are $\ge 0.4$. These denoising results reaffirm the effectiveness and generality of our method.

\begin{figure}[!htbp]
  \centering
  \includegraphics[width=0.8\linewidth]{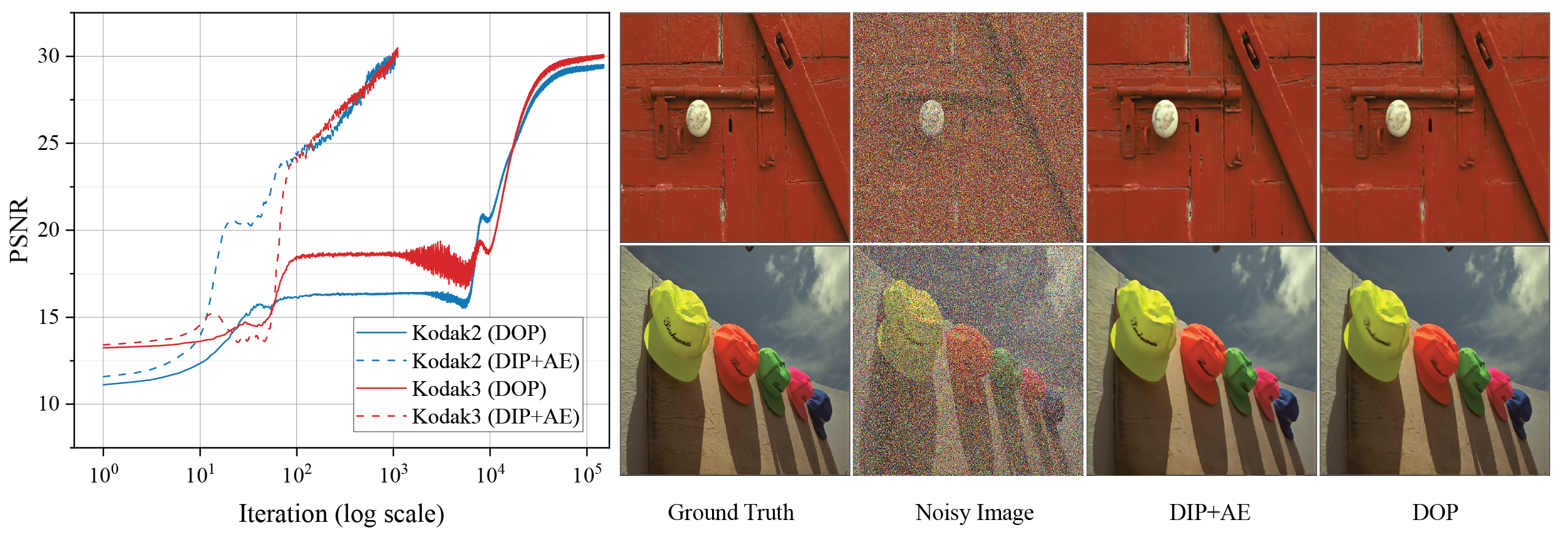}
  \caption{DIP+AE vs DOP. (left) The solid lines and dash lines respectively show the restoration performance for DOP and DIP+AE. (right) Visualizations for Kodak2 (first row) and Kodak3 (second row).}
  \label{fig:dop_vs_dip}
\end{figure}

Moreover, \cref{fig:dop_vs_dip} visualizes the reconstruction results of both DIP+AE and DOP. \cref{fig:dop_vs_dip} (left) shows that both DIP+AE and DOP attain similar performance in terms of PSNR while DIP+AE requires far fewer iterations and stops very early. \cref{fig:dop_vs_dip} (right) confirms that visually they also lead to similar reconstruction qualities, as there is almost no perceivable difference. 

\subsection{MRI reconstruction}\label{appendix:mri_reconstruction}
Here we provide the results for the complete set of random MRI samples that we experiment with. For the notations, \emph{ES} indicates the reconstruction quality detected by our method, \emph{Peak} denotes the peak quality that DD can achieve, and \emph{Overfitting} is the final reconstruction quality without ES. 
\begin{table*}[!htpb]
\caption{The experimental results of MRI reconstruction. }
\begin{center}
\setlength{\tabcolsep}{0.8mm}{
\begin{tabular}{c c|c c c c c c c}
\hline
& &\scriptsize{Sample 1} &\scriptsize{Sample 2} &\scriptsize{Sample 4} &\scriptsize{Sample 6} &\scriptsize{Sample 9} &\scriptsize{Sample 10} &\scriptsize{Sample 16}
\\
\hline
\multirow{3}{*}{\rotatebox{90}{\scriptsize{PSNR $\uparrow$ }}} 
& \scriptsize{ES}
& \tabincell{c}{\scriptsize{29.837} \tiny{(0.176)}} 
& \tabincell{c}{\scriptsize{30.077} \tiny{(0.231)}} 
& \tabincell{c}{\scriptsize{27.419} \tiny{(0.250)}} 
& \tabincell{c}{\scriptsize{28.799} \tiny{(0.111)}} 
& \tabincell{c}{\scriptsize{33.395} \tiny{(0.158)}} 
& \tabincell{c}{\scriptsize{27.907} \tiny{(0.124)}} 
& \tabincell{c}{\scriptsize{28.602} \tiny{(0.217)}} 
\\

& \scriptsize{Peak}
& \tabincell{c}{\scriptsize{31.792} \tiny{(0.295)}} 
& \tabincell{c}{\scriptsize{31.700} \tiny{(0.185)}} 
& \tabincell{c}{\scriptsize{28.465} \tiny{(0.282)}} 
& \tabincell{c}{\scriptsize{29.855} \tiny{(0.081)}} 
& \tabincell{c}{\scriptsize{34.444} \tiny{(0.265)}} 
& \tabincell{c}{\scriptsize{28.816} \tiny{(0.107)}} 
& \tabincell{c}{\scriptsize{30.494} \tiny{(0.381)}} 
\\

& \scriptsize{Overfitting}
& \tabincell{c}{\scriptsize{27.182} \tiny{(0.379)}} 
& \tabincell{c}{\scriptsize{26.710} \tiny{(0.449)}} 
& \tabincell{c}{\scriptsize{23.934} \tiny{(0.072)}} 
& \tabincell{c}{\scriptsize{25.631} \tiny{(0.191)}} 
& \tabincell{c}{\scriptsize{31.053} \tiny{(0.061)}} 
& \tabincell{c}{\scriptsize{25.545} \tiny{(0.182)}} 
& \tabincell{c}{\scriptsize{24.806} \tiny{(0.232)}} 
\\

\hline
\multirow{3}{*}{\rotatebox{90}{\scriptsize{SSIM $\uparrow$}}} 
& \scriptsize{ES}
& \tabincell{c}{\scriptsize{0.602} \tiny{(0.002)}} 
& \tabincell{c}{\scriptsize{0.609} \tiny{(0.000)}} 
& \tabincell{c}{\scriptsize{0.611} \tiny{(0.003)}} 
& \tabincell{c}{\scriptsize{0.612} \tiny{(0.005)}} 
& \tabincell{c}{\scriptsize{0.669} \tiny{(0.002)}} 
& \tabincell{c}{\scriptsize{0.620} \tiny{(0.008)}} 
& \tabincell{c}{\scriptsize{0.646} \tiny{(0.005)}} 
\\

& \scriptsize{Peak}
& \tabincell{c}{\scriptsize{0.642} \tiny{(0.006)}} 
& \tabincell{c}{\scriptsize{0.643} \tiny{(0.004)}} 
& \tabincell{c}{\scriptsize{0.658} \tiny{(0.003)}} 
& \tabincell{c}{\scriptsize{0.637} \tiny{(0.014)}} 
& \tabincell{c}{\scriptsize{0.689} \tiny{(0.008)}} 
& \tabincell{c}{\scriptsize{0.644} \tiny{(0.004)}} 
& \tabincell{c}{\scriptsize{0.671} \tiny{(0.003)}} 
\\

& \scriptsize{Overfitting}
& \tabincell{c}{\scriptsize{0.562} \tiny{(0.002)}} 
& \tabincell{c}{\scriptsize{0.571} \tiny{(0.002)}} 
& \tabincell{c}{\scriptsize{0.568} \tiny{(0.005)}} 
& \tabincell{c}{\scriptsize{0.507} \tiny{(0.012)}} 
& \tabincell{c}{\scriptsize{0.591} \tiny{(0.009)}} 
& \tabincell{c}{\scriptsize{0.554} \tiny{(0.004)}} 
& \tabincell{c}{\scriptsize{0.553} \tiny{(0.005)}} 
\\
\hline
\end{tabular}
}
\end{center}
\label{tab:mri_psnr_table}
\end{table*}
As can be seen from~\cref{tab:mri_psnr_table}, for all experimental samples, \emph{ES} and \emph{Peak} yield close performance in terms of both PSNR and SSIM, which indicates that our method can reliably detect the near-peak performance. On the other hand, the performance (both PSNR and SSIM) of \emph{ES} is uniformly better than that of \emph{overfitting}, often by a considerable margin, further confirming the effectiveness of our method.

\subsection{Image inpainting}\label{subsec:image_inpainting}
Image inpainting is another common IR task that SIDGPs have excelled in and hence popularly evaluated on; see, e.g., the DIP paper~\cite{ulyanov2018deep}. In this task, a clean image $x_0$ is contaminated by additive Gaussian noise $\eps$, and then only partially observed to yield the observation $y = (x_0 + \eps) \odot m$, where $m \in \{0, 1\}^{H \times W}$ is a binary mask and $\odot$ denotes the Hadamard pointwise product. Here both $y$ and $m$ are known, and the goal is to reconstruct $x_0$. We consider the natural formulation 
\begin{equation}
    E(x) = \norm{ (x-y) \odot m }_2^2. 
\end{equation}
We parametrize $x$ using the DIP model. The mask $m$ is generated according to an iid Bernoulli model, with a rate of $50\%$, i.e., $50\%$ of pixels not observed in expectation. The noise $\eps$ is set to the medium level.  

We also compare the performance of our method (DIP+AE) with the two competing methods: DIP+TV~\cite{sun2020solving,sun2021plug} and SGLD~\cite{cheng2019bayesian}. To ensure convergence, we run 
60K and 200K iterations for DIP+TV and SGLD, respectively, and report their final results. We repeat all experiments $3$ times and obtain the mean and standard deviation for each instance.

\cref{tab:image_inpainting} summarizes the results. Our method significantly outperforms the other two in terms of both PSNR and SSIM. It should be noted that here we test a medium noise level, rather than the very low noise levels experimented with in the DIP+TV and SGLD papers. Although in their original papers overfitting seems to be gone, here we see a strike-back with a different noise level. So the two competing methods are at best sensitive to hyperparameters, which are tricky to set.

\begin{table}[!htpb]
\begin{center}
\caption{DIP+AE, DIP+TA, and SGLD for image inpainting. The best PSNRs are colored as \textcolor{red}{red}; the best SSIMs are colored as \textcolor{blue}{blue}.}
\label{tab:image_inpainting}
\setlength{\tabcolsep}{1.5mm}{
\begin{tabular}{l c c c| c c c}
\hline

\multirow{2}{*}{\textit{}} & 
\multicolumn{3}{c}{{\scriptsize{PSNR $\uparrow$}}} &
\multicolumn{3}{c}{{\scriptsize{SSIM $\uparrow$}}}
\\
\hline
&
\scriptsize{DIP+AE} &
\scriptsize{DIP+TV} &
\scriptsize{SGLD} &
\scriptsize{DIP+AE} &
\scriptsize{DIP+TV} &
\scriptsize{SGLD} 
 \\

\hline
\scriptsize{Barbara}
&\tabincell{c}{\textcolor{red}{\scriptsize{21.878}} \tiny{(0.101)}}
&\tabincell{c}{\scriptsize{17.790} \tiny{(0.024)}}
&\tabincell{c}{\scriptsize{15.326} \tiny{(0.062)}}
&\tabincell{c}{\textcolor{blue}{\scriptsize{0.522}} \tiny{(0.010)}}
&\tabincell{c}{\scriptsize{0.259} \tiny{(0.000)}}
&\tabincell{c}{\scriptsize{0.259} \tiny{(0.003)}}
\\

\hline
\scriptsize{Boat}
&\tabincell{c}{\textcolor{red}{\scriptsize{23.268}} \tiny{(0.445)}}
&\tabincell{c}{\scriptsize{18.071} \tiny{(0.040)}}
&\tabincell{c}{\scriptsize{15.211} \tiny{(0.020)}}
&\tabincell{c}{\textcolor{blue}{\scriptsize{0.534}} \tiny{(0.010)}}
&\tabincell{c}{\scriptsize{0.259} \tiny{(0.001)}}
&\tabincell{c}{\scriptsize{0.227} \tiny{(0.001)}}
\\

\hline
\scriptsize{House}
&\tabincell{c}{\textcolor{red}{\scriptsize{28.883}} \tiny{(0.300)}}
&\tabincell{c}{\scriptsize{18.362} \tiny{(0.022)}}
&\tabincell{c}{\scriptsize{15.566} \tiny{(0.059)}}
&\tabincell{c}{\textcolor{blue}{\scriptsize{0.767}} \tiny{(0.025)}}
&\tabincell{c}{\scriptsize{0.157} \tiny{(0.000)}}
&\tabincell{c}{\scriptsize{0.171} \tiny{(0.002)}}
\\

\hline
\scriptsize{Lena}
&\tabincell{c}{\textcolor{red}{\scriptsize{25.052}} \tiny{(0.246)}}
&\tabincell{c}{\scriptsize{18.264} \tiny{(0.040)}}
&\tabincell{c}{\scriptsize{15.481} \tiny{(0.084)}}
&\tabincell{c}{\textcolor{blue}{\scriptsize{0.644}} \tiny{(0.004)}}
&\tabincell{c}{\scriptsize{0.218} \tiny{(0.001)}}
&\tabincell{c}{\scriptsize{0.204} \tiny{(0.003)}}
\\

\hline
\scriptsize{Peppers}
&\tabincell{c}{\textcolor{red}{\scriptsize{26.251}} \tiny{(0.187)}}
&\tabincell{c}{\scriptsize{18.400} \tiny{(0.022)}}
&\tabincell{c}{\scriptsize{15.610} \tiny{(0.036)}}
&\tabincell{c}{\textcolor{blue}{\scriptsize{0.738}} \tiny{(0.016)}}
&\tabincell{c}{\scriptsize{0.203} \tiny{(0.000)}}
&\tabincell{c}{\scriptsize{0.199} \tiny{(0.001)}}
\\

\hline
\scriptsize{C.man}
&\tabincell{c}{\textcolor{red}{\scriptsize{26.194}} \tiny{(0.423)}}
&\tabincell{c}{\scriptsize{18.571} \tiny{(0.049)}}
&\tabincell{c}{\scriptsize{15.861} \tiny{(0.031)}}
&\tabincell{c}{\textcolor{blue}{\scriptsize{0.732}} \tiny{(0.009)}}
&\tabincell{c}{\scriptsize{0.204} \tiny{(0.001)}}
&\tabincell{c}{\scriptsize{0.215} \tiny{(0.001)}}
\\

\hline
\scriptsize{Couple}
&\tabincell{c}{\textcolor{red}{\scriptsize{22.619}} \tiny{(0.154)}}
&\tabincell{c}{\scriptsize{18.115} \tiny{(0.007)}}
&\tabincell{c}{\scriptsize{15.313} \tiny{(0.062)}}
&\tabincell{c}{\textcolor{blue}{\scriptsize{0.512}} \tiny{(0.011)}}
&\tabincell{c}{\scriptsize{0.280} \tiny{(0.000)}}
&\tabincell{c}{\scriptsize{0.241} \tiny{(0.002)}}
\\

\hline
\scriptsize{Finger}
&\tabincell{c}{\textcolor{red}{\scriptsize{21.396}} \tiny{(0.119)}}
&\tabincell{c}{\scriptsize{17.714} \tiny{(0.024)}}
&\tabincell{c}{\scriptsize{15.150} \tiny{(0.027)}}
&\tabincell{c}{\textcolor{blue}{\scriptsize{0.795}} \tiny{(0.003)}}
&\tabincell{c}{\scriptsize{0.601} \tiny{(0.000)}}
&\tabincell{c}{\scriptsize{0.490} \tiny{(0.001)}}
\\

\hline
\scriptsize{Hill}
&\tabincell{c}{\textcolor{red}{\scriptsize{24.216}} \tiny{(0.254)}}
&\tabincell{c}{\scriptsize{18.274} \tiny{(0.017)}}
&\tabincell{c}{\scriptsize{15.514} \tiny{(0.107)}}
&\tabincell{c}{\textcolor{blue}{\scriptsize{0.518}} \tiny{(0.009)}}
&\tabincell{c}{\scriptsize{0.242} \tiny{(0.000)}}
&\tabincell{c}{\scriptsize{0.214} \tiny{(0.003)}}
\\

\hline
\scriptsize{Man}
&\tabincell{c}{\textcolor{red}{\scriptsize{23.687}} \tiny{(0.302)}}
&\tabincell{c}{\scriptsize{18.159} \tiny{(0.022)}}
&\tabincell{c}{\scriptsize{15.394} \tiny{(0.109)}}
&\tabincell{c}{\textcolor{blue}{\scriptsize{0.532}} \tiny{(0.007)}}
&\tabincell{c}{\scriptsize{0.252} \tiny{(0.001)}}
&\tabincell{c}{\scriptsize{0.222} \tiny{(0.004)}}
\\

\hline
\scriptsize{Montage}
&\tabincell{c}{\textcolor{red}{\scriptsize{27.290}} \tiny{(0.282)}}
&\tabincell{c}{\scriptsize{19.005} \tiny{(0.022)}}
&\tabincell{c}{\scriptsize{16.334} \tiny{(0.017)}}
&\tabincell{c}{\textcolor{blue}{\scriptsize{0.799}} \tiny{(0.007)}}
&\tabincell{c}{\scriptsize{0.193} \tiny{(0.000)}}
&\tabincell{c}{\scriptsize{0.221} \tiny{(0.001)}}
\\







\bottomrule
\end{tabular}
}
\end{center}
\end{table}

\subsection{Performance on clean images}
\quad \label{subsec:clean}
For SIDGPs, when there is no noise, the target clean image is a global optimizer to \cref{eq:inv_obj_2}. So there is no overfitting issue in these scenarios, and ES is not strictly necessary. But, in practice, one does not know if noise is present apriori, and finite termination has to be made. In this section, we experiment with ``denoising'' \emph{clean} images with DIP. The setup is exactly as that of our typical denoising, except that here we do not report the PSNR gap, as whenever one makes a stop after finite iterations, the theoretical PNSR gap is infinity. We report the absolute PSNR detected by our method instead; for most applications, PNSR greater than $30$ is good enough for practical purposes. 

The AE error curve tends to fluctuate when the quality is already high. To improve the detection performance, we find that comparing running average to determine ES points performs better than the stopping criterion described in \cref{alg:framework}. We use this slightly modified version here; we leave reconciling the two versions as future work---we suspect that this modified version will likely improve our previous detection performance. 

\cref{tab:clean_setting} shows the preliminary results. Except for the challenging case \texttt{Baboon}, the PSNR scores are near $30$ or above. So our method is performing reasonable detection. We suspect using more advanced smoothing techniques such that Gaussian smoothing can suppress the fluctuation better and hence lead to better performance; we leave this as future work. 
\begin{table}[!htpb]
\begin{center}
\caption{Performance of DIP+AE on denoising clean images.}
\label{tab:clean_setting}
\setlength{\tabcolsep}{7.5mm}{
\begin{tabular}{l c c}
\hline


&
{\scriptsize{PSNR $\uparrow$}}&
\scriptsize{SSIM $\uparrow$}\\

\hline
\scriptsize{House}
&\tabincell{c}{\scriptsize{36.569} }
&\tabincell{c}{\scriptsize{0.921} }
\\

\hline
\scriptsize{Peppers}
&\tabincell{c}{\scriptsize{30.407} }
&\tabincell{c}{\scriptsize{0.797} }
\\

\hline
\scriptsize{Lena}
&\tabincell{c}{\scriptsize{31.927} }
&\tabincell{c}{\scriptsize{0.857} }
\\

\hline
\scriptsize{Baboon}
&\tabincell{c}{\scriptsize{20.186} }
&\tabincell{c}{\scriptsize{0.423} }
\\

\hline
\scriptsize{F16}
&\tabincell{c}{\scriptsize{33.065} }
&\tabincell{c}{\scriptsize{0.911} }
\\

\hline
\scriptsize{Kodak1}
&\tabincell{c}{\scriptsize{29.243} }
&\tabincell{c}{\scriptsize{0.852} }
\\

\hline
\scriptsize{Kodak2}
&\tabincell{c}{\scriptsize{31.064} }
&\tabincell{c}{\scriptsize{0.826} }
\\

\hline
\scriptsize{Kodak3}
&\tabincell{c}{\scriptsize{30.155} }
&\tabincell{c}{\scriptsize{0.861} }
\\

\hline
\scriptsize{Kodak12}
&\tabincell{c}{\scriptsize{31.757} }
&\tabincell{c}{\scriptsize{0.851} }
\\

\hline
\end{tabular}
}
\end{center}
\vspace{-2em}
\end{table}

\begin{table*}[t]
\caption{The performance gaps of BRISQUE~\cite{mittal2012no}, NIQE~\cite{mittal2012making}, NIMA~\cite{talebi2018nima}, and DIP+AE on shot and speckle noises.  For NIMA, we report both technical quality assessment (the number before ``/'') and aesthetic assessment (the number after ``/''). The best PSNR gaps are colored as \textcolor{red}{red}; the best SSIM gaps are colored as \textcolor{blue}{blue}.}
\vspace{-1em}
\label{tab:other_metric_part2}
\begin{center}
\setlength{\tabcolsep}{0.35mm}{
\begin{tabular}{l c c c c| c c c c|| c c c c| c c c c}
& 
\multicolumn{8}{c}{\scriptsize{Shot noise}} &
\multicolumn{8}{c}{\scriptsize{Speckle noise}}
\\
\hline
&\multicolumn{4}{c|}{\tiny{Gap in PSNR $\downarrow$}} 
&\multicolumn{4}{c||}{\tiny{Gap in SSIM $\downarrow$}} 
&\multicolumn{4}{c|}{\tiny{Gap in PSNR $\downarrow$}} 
&\multicolumn{4}{c}{\tiny{Gap in SSIM $\downarrow$}}
\\
\hline
& \tiny{BRISQUE}
& \tiny{NIQE}
& \tiny{NIMA}
& \tiny{DIP+AE}

& \tiny{BRISQUE}
& \tiny{NIQE}
& \tiny{NIMA}
& \tiny{DIP+AE}

& \tiny{BRISQUE}
& \tiny{NIQE}
& \tiny{NIMA}
& \tiny{DIP+AE}

& \tiny{BRISQUE}
& \tiny{NIQE}
& \tiny{NIMA}
& \tiny{DIP+AE}
\\
\hline
\tiny{House}
&\tiny{6.713}
&\tiny{8.629}
&\tiny{10.873/0.662}
&\tiny{\textcolor{red}{0.294}}

&\tiny{0.389}
&\tiny{0.491}
&\tiny{0.598/0.024}
&\tiny{\textcolor{blue}{0.002}}

&\tiny{9.848}
&\tiny{8.970}
&\tiny{12.879/\textcolor{red}{1.394}}
&\tiny{1.847}

&\tiny{0.457}
&\tiny{0.424}
&\tiny{0.591/0.027}
&\tiny{\textcolor{blue}{0.010}}
\\
\hline
\tiny{Peppers}
&\tiny{5.538}
&\tiny{5.975}
&\tiny{1.863/6.013}
&\tiny{\textcolor{red}{0.417}}

&\tiny{0.267}
&\tiny{0.289}
&\tiny{0.128/0.295}
&\tiny{\textcolor{blue}{0.018}}

&\tiny{6.414}
&\tiny{6.085}
&\tiny{8.987/4.861}
&\tiny{\textcolor{red}{0.311}}

&\tiny{0.233}
&\tiny{0.227}
&\tiny{0.316/0.193}
&\tiny{\textcolor{blue}{0.011}}
\\

\hline
\tiny{Lena}
&\tiny{7.976}
&\tiny{6.191}
&\tiny{9.697/1.545}
&\tiny{\textcolor{red}{1.281}}

&\tiny{0.448}
&\tiny{0.375}
&\tiny{0.516/0.075}
&\tiny{\textcolor{blue}{0.020}}

&\tiny{8.797}
&\tiny{5.133}
&\tiny{9.352/0.912}
&\tiny{\textcolor{red}{0.445}}

&\tiny{0.402}
&\tiny{0.240}
&\tiny{0.436/0.039}
&\tiny{\textcolor{blue}{0.013}}
\\

\hline
\tiny{Baboon}
&\tiny{\textcolor{red}{0.508}}
&\tiny{0.562}
&\tiny{2.767/3.061}
&\tiny{1.920}

&\tiny{0.026}
&\tiny{\textcolor{blue}{0.009}}
&\tiny{0.391/0.404}
&\tiny{0.284}

&\tiny{\textcolor{red}{0.387}}
&\tiny{1.507}
&\tiny{1.178/2.063}
&\tiny{2.318}

&\tiny{\textcolor{blue}{0.027}}
&\tiny{0.053}
&\tiny{0.144/0.231}
&\tiny{0.314}
\\

\hline
\tiny{F16}
&\tiny{5.329}
&\tiny{8.448}
&\tiny{\textcolor{red}{0.938}/6.404}
&\tiny{1.016}

&\tiny{0.403}
&\tiny{0.555}
&\tiny{\textcolor{blue}{0.011}/0.168}
&\tiny{0.013}

&\tiny{6.760}
&\tiny{7.136}
&\tiny{\textcolor{red}{0.418}/7.757}
&\tiny{\textcolor{red}{0.418}}

&\tiny{0.488}
&\tiny{0.498}
&\tiny{\textcolor{blue}{0.001}/0.229}
&\tiny{0.021}
\\

\hline
\tiny{Kodak1}
&\tiny{3.099}
&\tiny{3.741}
&\tiny{3.287/5.387}
&\tiny{\textcolor{red}{0.953}}

&\tiny{0.086}
&\tiny{0.118}
&\tiny{0.321/0.485}
&\tiny{\textcolor{blue}{0.059}}

&\tiny{1.363}
&\tiny{4.332}
&\tiny{3.215/6.871}
&\tiny{\textcolor{red}{0.719}}

&\tiny{\textcolor{blue}{0.013}}
&\tiny{0.106}
&\tiny{0.258/0.556}
&\tiny{0.056}
\\

\hline
\tiny{Kodak2}
&\tiny{15.693}
&\tiny{9.055}
&\tiny{10.208/1.071}
&\tiny{\textcolor{red}{0.154}}

&\tiny{0.433}
&\tiny{0.269}
&\tiny{0.435/0.025}
&\tiny{\textcolor{blue}{0.006}}

&\tiny{9.755}
&\tiny{7.921}
&\tiny{8.169/0.634}
&\tiny{\textcolor{red}{0.340}}

&\tiny{0.272}
&\tiny{0.177}
&\tiny{0.158/\textcolor{blue}{0.013}}
&\tiny{\textcolor{blue}{0.013}}
\\

\hline
\tiny{Kodak3}
&\tiny{8.429}
&\tiny{8.546}
&\tiny{2.501/21.796}
&\tiny{\textcolor{red}{0.895}}

&\tiny{0.429}
&\tiny{0.427}
&\tiny{0.092/0.619}
&\tiny{\textcolor{blue}{0.006}}

&\tiny{6.484}
&\tiny{6.509}
&\tiny{13.457/13.130}
&\tiny{\textcolor{red}{1.183}}

&\tiny{0.236}
&\tiny{0.239}
&\tiny{0.584/0.263}
&\tiny{\textcolor{blue}{0.016}}
\\

\hline
\tiny{Kodak12}
&\tiny{6.009}
&\tiny{8.941}
&\tiny{6.746/20.054}
&\tiny{\textcolor{red}{2.118}}

&\tiny{0.401}
&\tiny{0.490}
&\tiny{0.422/0.704}
&\tiny{\textcolor{blue}{0.021}}

&\tiny{8.545}
&\tiny{8.204}
&\tiny{5.398/0.703}
&\tiny{\textcolor{red}{0.562}}

&\tiny{0.459}
&\tiny{0.450}
&\tiny{0.343/0.011}
&\tiny{\textcolor{blue}{0.010}}
\\
\hline
\end{tabular}
}
\end{center}
\end{table*}

\subsection{Bell-shape examples under different learning rate}

\begin{figure}[t ]
  \centering
  \includegraphics[width=0.8\linewidth]{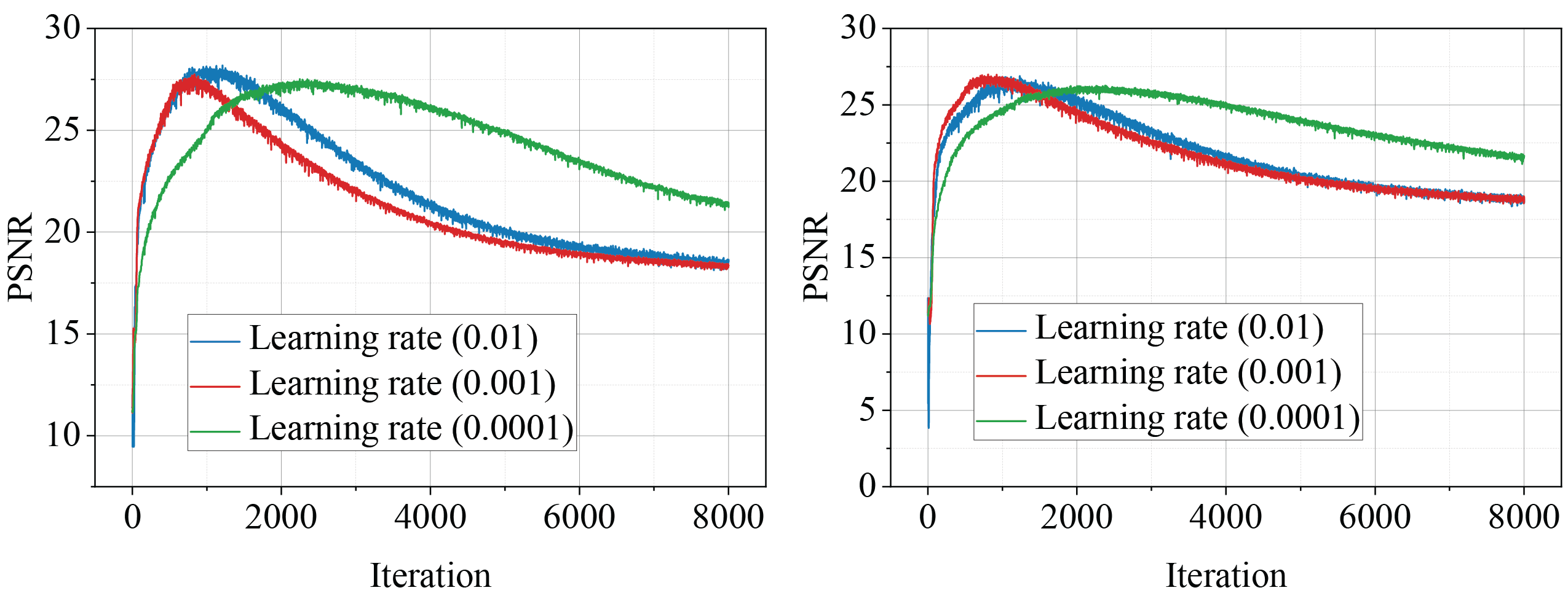}
 
  \caption{The PSNR curves of DIP under different learning rates. (left) the PSNR curves for F16; (right) the PSNR curves for Peppers.}
  \label{fig:different_LR}
\end{figure}

As we shown in~\cref{tab:diff_LR}, our ES detection method is stable in terms of different learning rates of DIP. Here we further demonstrate that the bell-shape of PSNR curve of DIP is holds under different learning rates. We randomly select two images---F16 and Peppers---and visualize their PSNR curves under different learning rates \{0.01, 0.001, 0.0001\} in~\cref{fig:different_LR}. We can observe that different rates would perturb the curves but would not distort the overall bell shape. 

\subsection{Analysis of failure mode}
\begin{figure}[t]
  \centering
  \includegraphics[width=0.8\linewidth]{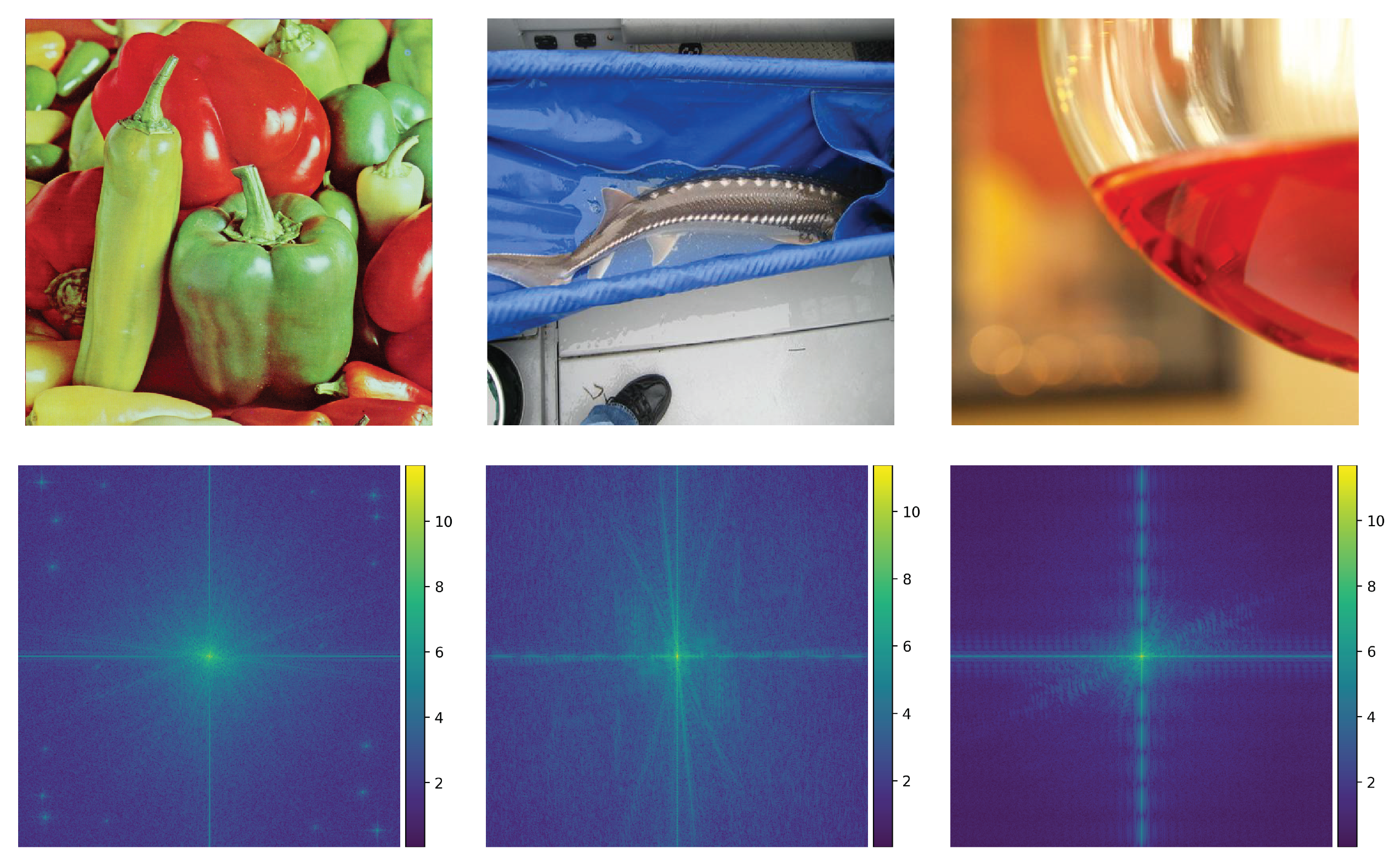}
  \caption{Positive images and their spectra.}
  \label{fig:good_img}
\end{figure}

\begin{figure}[!htbp]
  \centering
  \includegraphics[width=0.8\linewidth]{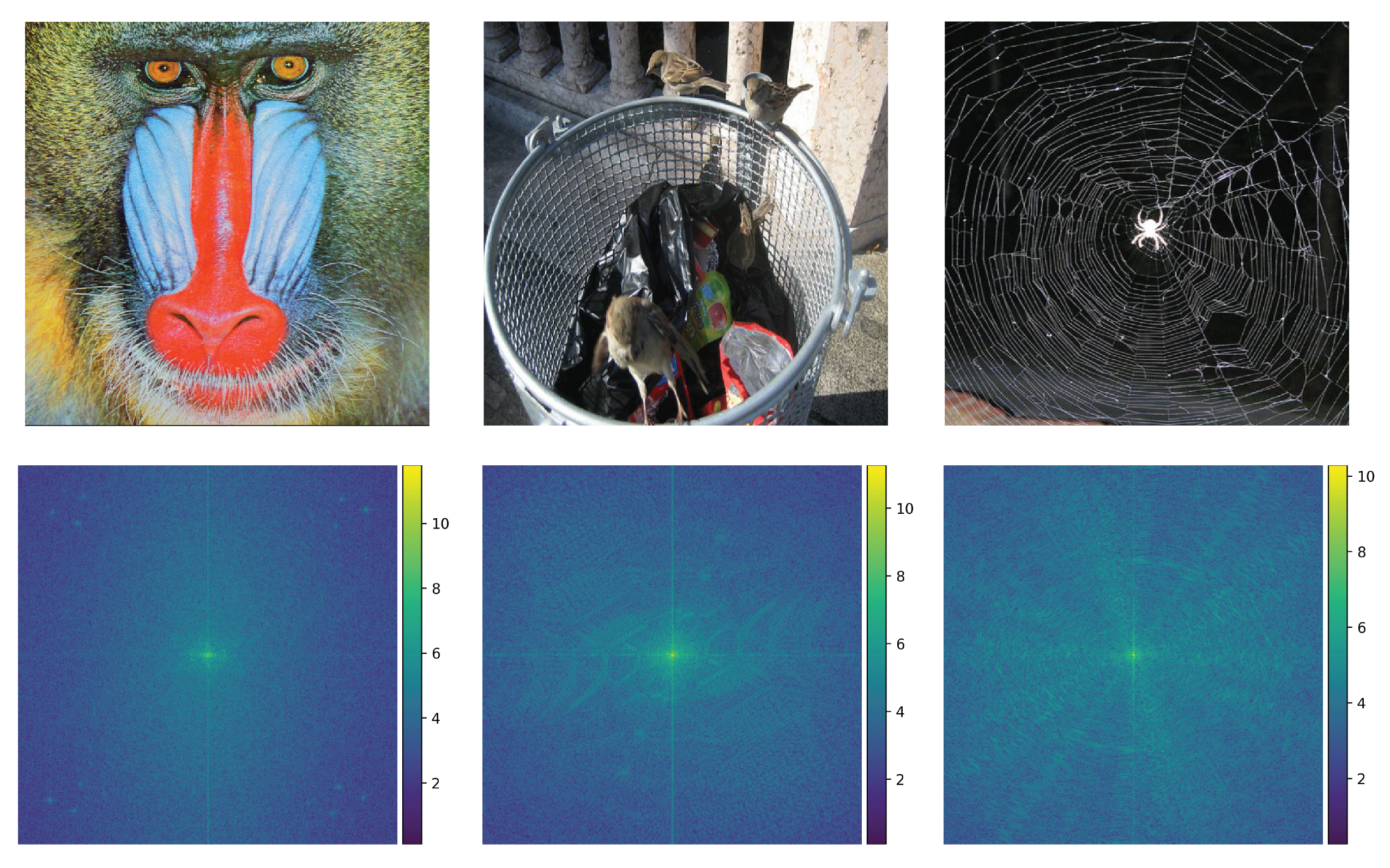}
  \caption{Negative images and their spectra.}
  \label{fig:bad_img}
\end{figure}

To qualitatively understand the failing cases, we select $3$ positive images that enjoy consistent good detection and $3$ negative images that see consistent failure, and visualize their Fourier spectra in \cref{fig:good_img,fig:bad_img}. For better visualization, we take the $x \mapsto \log(1+x)$ transform of the Fourier magnitudes, as is commonly done in image processing. Visually, the positive examples can be well characterized as being piecewise smooth, and the negative ones invariably contain fine details that correspond to high frequency components. Indeed, the positive spectra are concentrated in the low-frequency bands, whereas the negative spectra are much more scattered into high-frequency bands. We leave a more quantitative analysis of this as future work. 

\subsection{ES criterion based on other principles?}
During the peer review, one reviewer kindly pointed to us the possibility of formulating ES criteria based on the \emph{whiteness} or \emph{discrepancy} principles in image denoising~\cite{LanzaEtAl2018Whiteness,AlmansaEtAl2007TV}. In this section, we briefly discuss the possibility of implementing them. We want to quickly remind that we target practical denoising, and so assume very little knowledge about noise types, levels, or whatsoever. Particularly, the noise level $\sigma^2$ is unknown to us and possibly also hard to estimate due to the generality we strive for. Also, to avoid confusion, we will switch our notations slightly here. 

Let $\mb X \in \R^{n \times n}$ be the target image, and $\mb Y = f\paren{\mb X} + \mb W$ be the noisy measurement, where $\mb W \in \R^{n \times n}$ is white noise with variance level $\sigma^2 < +\infty$, i.e. 
\begin{itemize}
    \item $\bb E\, w_{ij} = 0$ and $\bb E\, w_{ij}^2 = \sigma^2\quad \forall\; i, j$; 
    \item any pair of distinct elements in $\mb W$ are \emph{uncorrelated}: for two different (in location) elements $w, w'$ of $\mb W$,  $\expect{(w-\bb E\, w)(w'- \bb E\, w')} = 0 \Longrightarrow \expect{ww'} = 0$ (the implication uses the 1st property). 
\end{itemize}
The 2nd property has an interesting implication: 
\begin{align}
    \expect{\mb W \star \mb W} = n^2 \sigma^2 \delta_{n \times n} = 
    \begin{bmatrix}
        n^2 \sigma^2 & \mb 0^\T_{n-1} \\
        \mb 0_{n-1} &  \mb 0_{(n-1) \times (n-1)} 
    \end{bmatrix}, 
\end{align}
where $\star$ denotes the 2D cross-correlation (for convenience, we assume the circular version), and $\delta_{n\times n}$ is the 2D delta function (we assume the top-left corner corresponds to no-shift alignment). Denote our estimated image as $\wh{\mb X}$. If we have perfect recovery, then 
\begin{align}
    \mb Y - f\paren{\wh{\mb X}} = \mb Y - f\paren{\mb X} = \mb W, 
\end{align}
and so 
\begin{align}
    \expect{\paren{\mb Y - f\paren{\wh{\mb X}}} \star \paren{\mb Y - f\paren{\wh{\mb X}}}} = \expect{\mb W \star \mb W} = n^2 \sigma^2 \delta_{n \times n}. 
\end{align}
This is the \emph{whiteness principle} in~\cite{LanzaEtAl2018Whiteness}\footnote{Our presentation here does not use the discrete-time stochastic process language as in the original paper---which seems overly technical than necessary, but they are equivalent.}: in particular, except for the top-left corner, all other elements should be zero. 

In practice, we are not able to take the expectation as we only observe one realization. But note that except for the perfectly aligned case which produces $n^2 \sigma^2$, each element of $\frac{1}{n^2}\mb W \star \mb W$ can be approximated by $\mc N\paren{0, \frac{\sigma^4}{n^2}}$ due to central limit theorem when $n^2$ is large---this is valid for images. So typical element of $\mb W \star \mb W$ should lie in the range 
\begin{align}
    [-c n \sigma^2, c n\sigma^2], 
\end{align}
where $c>1$ is a large enough constant, say $5$. 

This is explicitly used as a constraint in~\cite{LanzaEtAl2018Whiteness} to improve image restoration quality. If the variance level $\sigma^2$ is known, we may be able to use distance to this set as a measure of image quality. However, for our applications, we do not assume known $\sigma^2$, and also the constant $c$ chosen can impact the result also. 

A more reasonable metric for us would be the quantity 
\begin{align}
    \norm{(\mb W \star \mb W)_{-0}}, 
\end{align}
i.e., the norm of the $\mb W \star \mb W$ with the top-left corner removed, which ideally should be as sufficiently small. To be more explicit, 
\begin{align}
    \norm{\paren{\paren{\mb Y - f\paren{\wh{\mb X}}} \star \paren{\mb Y - f\paren{\wh{\mb X}}}}_{-0}}. 
\end{align} 
But obviously the minimum is achieved when we overfit the noisy image $\mb Y$. 

The \emph{discrepancy principle} (or local constraint) in~\cite{AlmansaEtAl2007TV} is a refinement to the obvious constraint 
\begin{align} 
    \frac{1}{n^2}\norm{f\paren{\wh{\mb X}} - \mb Y}_F^2 \le \sigma^2 
\end{align}
for a good estimate $\wh{\mb X}$ to satisfy. This only enforces that \emph{globally} the noise level of the residual matches the known noise level, but does not ensure \emph{uniformly}. To ensure the latter, a natural idea is to enforce the noise level consistently everywhere locally: 
\begin{align}
G \ast \paren{f\paren{\wh{\mb X}} - \mb Y}^2_{i, j} \le \sigma^2 \quad \forall\; i, j. 
\end{align} 
Here, $G$ is a Gaussian filter of appropriate size, and the left side of the equality is the (Gaussian) weighted mean variance around the pixel location $(i, j)$. 

Similar to the situation for the whiteness principle, if the variance level $\sigma^2$ is known, we may also use the distance to this set as a metric to measure the reconstruction quality. When $\sigma^2$ is unknown, it is unclear how to make easy modification to do this, unlike the case of the whiteness principle above.

\end{document}